\documentclass[journal]{IEEEtran}
\usepackage{amsmath,amsfonts}
\usepackage{algorithmic}
\usepackage{algorithm}
\usepackage{array}
\usepackage[caption=false,font=normalsize,labelfont=rm,textfont=rm]{subfig}
\usepackage{textcomp}
\usepackage{stfloats}
\usepackage{url}
\usepackage{verbatim}
\usepackage{graphicx}
\usepackage{cite}
\usepackage{pifont}
\usepackage{tikz}
\usetikzlibrary{spy}
\usepackage{xspace}
\usepackage{booktabs}
\usepackage{multirow}
\usepackage{float}
\usepackage{overpic}
\usepackage[pagebackref,breaklinks,colorlinks]{hyperref}
\usepackage{amssymb}
\usepackage[capitalize]{cleveref}
\usepackage{bbding}
\usepackage{caption}
\usepackage{bm}
\usepackage{tablefootnote}

\crefname{section}{Sec.}{Secs.}
\Crefname{section}{Section}{Sections}
\Crefname{table}{Table}{Tables}
\crefname{table}{Tab.}{Tabs.}

\newcommand{\ie}{{\emph{i.e.}}\xspace}

\newcommand{\eg}{{\emph{e.g.}}\xspace}

\newcommand{\etal}{{\emph{et al.}}}

\newcommand{\myname}[0]{St-EDNet}
\newcommand{\mydata}[0]{StEIC}

\begin{document}

\title{Learning Parallax for Stereo Event-based \\ Motion Deblurring}

\author{Mingyuan Lin, Chi Zhang, Chu He, and Lei Yu
\thanks{M. Lin, C. Zhang, C. He, and L. Yu are with the School of Electronic Information, Wuhan University, Wuhan 430072, China. E-mail: \{linmingyuan, zhangchi1, chuhe, ly.wd\}@whu.edu.cn.}
\thanks{The research was partially supported by the National Natural Science
Foundation of China under Grants 62271354 and 61871297.}
\thanks{Corresponding authors: Chu He and Lei Yu.}
}

\markboth{SUBMISSION TO IEEE Transactions on Circuits and Systems for Video Technology}%
{Shell \MakeLowercase{\textit{et al.}}: A Sample Article Using IEEEtran.cls for IEEE Journals}


\maketitle

\begin{abstract}
Due to the extremely low latency, events have been recently exploited to supplement lost information for motion deblurring. Existing approaches 
largely rely on the perfect pixel-wise alignment between intensity images and events, which is not always fulfilled in the real world.
To tackle this problem, we propose a novel coarse-to-fine framework, named NETwork of Event-based motion Deblurring with STereo event and intensity cameras (\myname), to recover high-quality images directly from the misaligned inputs, consisting of a single blurry image and the concurrent event streams. Specifically, the coarse spatial alignment of the blurry image and the event streams is first implemented with a cross-modal stereo matching module without the need for ground-truth depths. Then, a dual-feature embedding architecture is proposed to gradually build the fine bidirectional association of the coarsely aligned data and reconstruct the sequence of the latent sharp images. Furthermore, we build a new dataset with STereo Event and Intensity Cameras (\mydata), containing real-world events, intensity images, and dense disparity maps. Experiments on real-world datasets demonstrate the superiority of the proposed network over state-of-the-art methods. The code and dataset are available at \url{https://mingyuan-lin.github.io/St-ED_web/}.
\end{abstract}

\begin{IEEEkeywords}
Motion Deblurring, Stereo Matching, Event Camera.
\end{IEEEkeywords}

\section{Introduction}
\IEEEPARstart{M}{OTION} blur is a common image degradation in fast-moving photography. Restoring sharp textures from a single blurry image is highly ill-posed due to missing information in terms of texture erasures and motion ambiguities~\cite{jin2018learning,purohit2019bringing,rengarajan2020photosequencing}.
The event camera can supplement the missing information thanks to its low latency and high dynamics \cite{lichtsteiner2008128,brandli2014240,gallego2020event}, which helps yield reliable reconstruction results. Recently, many event-based motion deblurring approaches have been proposed that commonly rely on the assumption of per-pixel alignments between blurry images and event streams \cite{pan2019bringing,lin2020learning,stoffregen2020reducing,wang2020event,xu2021motion,song2022cir}. However, the above assumption is not always fulfilled in real-world applications, \eg, the stereo event and intensity camera setup~\cite{gehrig2021dsec,cho2022event}, leading to severe degradation of the motion deblurring performance.

Although spatially aligned events and intensity frames can be captured by a shared sensor in an event-intensity camera, \eg, the DAVIS event camera~\cite{brandli2014240}, the low resolution of such cameras hinders the popularization of existing methods in practical applications. Therefore, the association and fusion of multi-sensor and multi-modal data in the stereo event and intensity camera setup are more practical but more challenging due to the existence of parallax between the event camera and the intensity camera, as shown in Fig. \ref{fig:first}. In this case, a pixel-level alignment is essential but ill-posed for event-based motion deblurring approaches since motion brings coupled burdens for multi-modal correspondence, \ie, {\it blurry effects} and {\it dynamic scene depth}.

\def\ssxxsone{(0.1, -.4)} 
\def\ssyysone{(0.95,0.56)} 
\def\ssxxstwo{(-0.37,0.1)} 
\def\ssyystwo{(0.95,0.56)} 
\def\ssizz{0.8cm} 
\def\sswidth{0.245\linewidth} 
\def\ssmag{2}
\def\scc{(-1.4,0.92)}
\def\sccone{(-1.4,0.92)}

\begin{figure}
	\centering
    \begin{minipage}[b]{\linewidth}
	    \centering
	    \subfloat[Stereo event and intensity camera setup.]{\includegraphics[width=\linewidth]{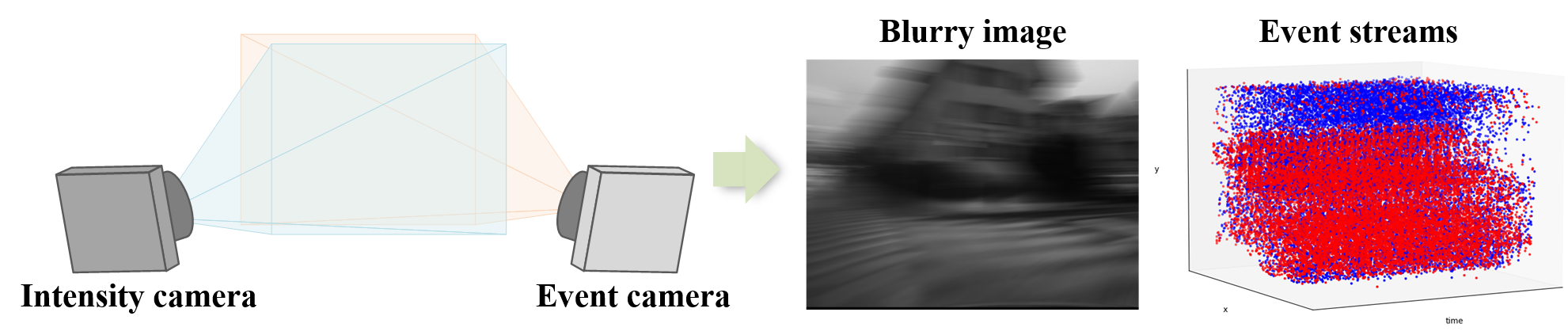}}
    \end{minipage}\\

    \begin{minipage}[b]{\linewidth}
	    \centering
	    \subfloat[Comparison of motion deblurring results of three methods.]{
    	\begin{tabular}{c c c c}
    	\hspace{-3mm}\begin{tikzpicture}[spy using outlines={green,magnification=\ssmag,size=\ssizz},inner sep=0]
    		\node {\includegraphics[width=\sswidth]{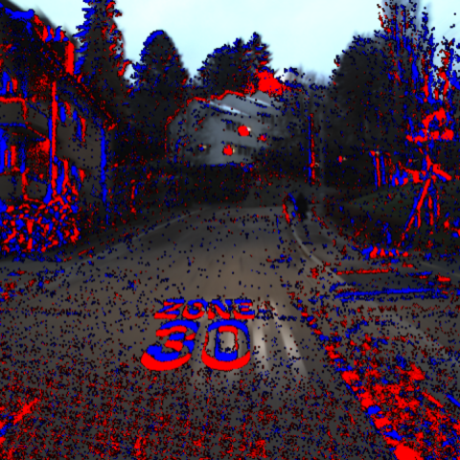}};
    	\end{tikzpicture}& \hspace{-4.8mm}
    		\begin{tikzpicture}[spy using outlines={green,magnification=\ssmag,size=\ssizz},inner sep=0]
    		\node {\includegraphics[width=\sswidth]{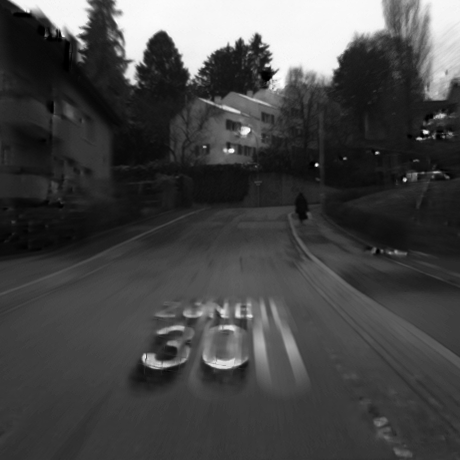}};
    	\end{tikzpicture}& \hspace{-4.8mm}
    		\begin{tikzpicture}[spy using outlines={green,magnification=\ssmag,size=\ssizz},inner sep=0]
    		\node {\includegraphics[width=\sswidth]{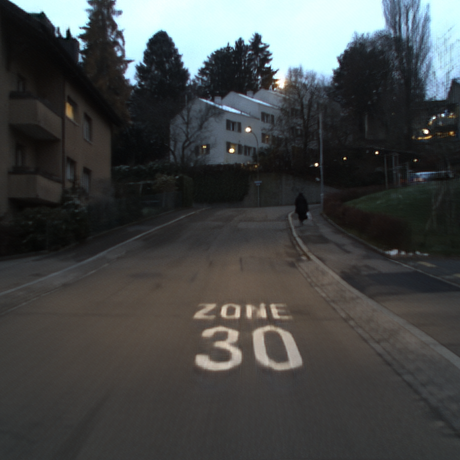}};
    	\end{tikzpicture}& \hspace{-4.8mm}
    	\begin{tikzpicture}[spy using outlines={green,magnification=\ssmag,size=\ssizz},inner sep=0]
    		\node {\includegraphics[width=\sswidth]{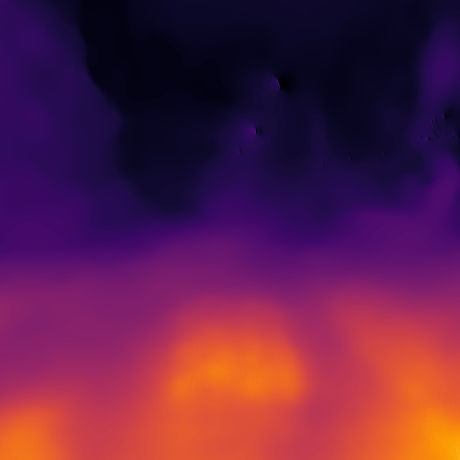}};
    	\end{tikzpicture}\vspace{-0.8mm}\\
    	\hspace{-3mm}\begin{tikzpicture}[spy using outlines={green,magnification=\ssmag,size=\ssizz},inner sep=0]
    		\node {\includegraphics[width=\sswidth]{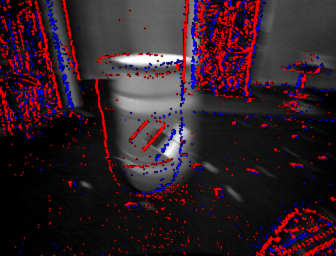}};
    	\end{tikzpicture}& \hspace{-4.8mm}
    	\begin{tikzpicture}[spy using outlines={green,magnification=\ssmag,size=\ssizz},inner sep=0]
    		\node {\includegraphics[width=\sswidth]{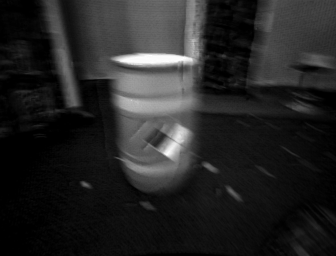}};
    	\end{tikzpicture}& \hspace{-4.8mm}
    		\begin{tikzpicture}[spy using outlines={green,magnification=\ssmag,size=\ssizz},inner sep=0]
    		\node {\includegraphics[width=\sswidth]{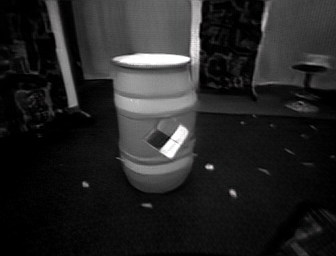}};
    	\end{tikzpicture}& \hspace{-4.8mm}
    		\begin{tikzpicture}[spy using outlines={green,magnification=\ssmag,size=\ssizz},inner sep=0]
    		\node {\includegraphics[width=\sswidth]{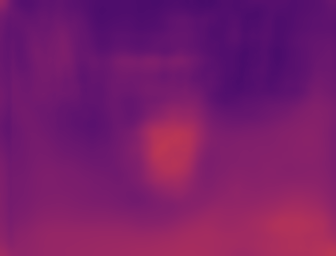}};
    	\end{tikzpicture}\vspace{-0.2mm}\\
    	\hspace{-3mm}\footnotesize{Inputs} & \hspace{-3mm}\footnotesize{LEDVDI \cite{lin2020learning}} & \hspace{-3mm}\footnotesize{Our image} & \hspace{-3mm}\footnotesize{Our disparity}
    	\end{tabular}}
    \end{minipage}

	\caption{Illustrative examples of the impact of the misaligned event and frame data from the \textit{DSEC} and \textit{MVSEC} datasets. Our \myname\ generates fewer artifacts and achieves the best visualization performance.}
	\label{fig:first}
	\vspace{-4mm}
\end{figure}

\begin{itemize}
    \item {\bf Blurry Effects.} When perceiving the same light field, events and intensities exhibit in different modalities but implicitly share common structures, enabling pixel-level correspondence \cite{zhangdata}. However, such correspondence becomes weak and ambiguous when intensity images appear blurry. 
    \item {\bf Dynamic Scene Depth.} Even though one can apply calibration \cite{huang2021dynamic,muglikar2021calibrate} and stereo rectification between the stereo event and intensity pairs as a pre-processing step to alleviate parallax \cite{heyden2005multiple,tulyakov2021time}, the resulted homography can not achieve the pixel-level correspondence since the scene depth is a varying and commonly unknown prior.
\end{itemize}

Therefore, this paper proposes a novel coarse-to-fine framework, named NETwork of Event-based motion Deblurring with STereo event and intensity cameras (\myname), to effectively dig out and aggregate the information of the input single blurry image and the corresponding event streams (alignment is not guaranteed), and output a sequence of sharp images and a disparity map. To relieve the above burdens, we first build a coarse stereo matching module, DispNet, to estimate the disparity and achieve a coarse association between the blurry image and the event streams. An alignment refinement and deblurring module named DblrNet receives the coarsely aligned data and generates a sequence of deblurred images. As the core architecture of DblrNet, the cascaded Disparity-guided Dual Feature Embedding (DDFE) modules are proposed to reconstruct fine correspondence between the coarsely aligned blurry image and events and further integrate and enhance the multi-modal information at the feature domain in a gradual manner. In summary, \myname\ can reconstruct clear images and estimate a disparity map simultaneously.



For training, the recently available datasets with the stereo event and intensity cameras, \eg, the Multi Vehicle Stereo Event Camera (\textit{MVSEC}) dataset \cite{zhu2018multivehicle} and the Stereo Event Camera dataset for Driving Scenarios (\textit{DSEC}) \cite{gehrig2021dsec}, due to the provided low frame rate images and sparse ground-truth depth maps by LiDAR, are not well suited for the motion deblurring task. It motivates us to build a stereo hybrid camera system and capture a new dataset with STereo Event and Intensity Cameras (\textit{\mydata}), which contains real-world events, intensity images, and dense ground-truth depth maps under various scenes, facilitating future research.

The main contributions of our work are five folds:
\begin{itemize}
    \item We propose the coarse-to-fine framework \myname\ to generate a sequence of clear images and estimate a disparity map simultaneously from a single blurry image and the concurrent event streams.
    \item We propose the Disparity-guided Dual Feature Embedding (DDFE) module to gradually supplement and enhance high-dimensional information with bidirectional disparities between different views at the feature level.
    \item We establish a real-world dataset with STereo Event and Intensity Cameras (\textit{\mydata}), which contains various scenes and diverse motions to facilitate future research. The code and dataset are available at \url{https://mingyuan-lin.github.io/St-ED_web/}.
\end{itemize}

\section{Related Work}
\subsection{Motion Deblurring}
Even though motion blur throws away intra-frame information by averaging over the exposure time interval, it inherently embeds both motions and textures of the moving objects, and thus enables the possibility of motion deblurring for the reconstruction of the dynamic scene behind a blurred photograph \cite{jin2018learning,jin2019learning,purohit2019bringing,bai2019single}. Motion is an essential clue to tackle deblurring tasks, and one can ease the deblurring task by assuming spatially uniform motion, degrading the blurring process to the convolution of blur kernel with the sharp image~\cite{krishnan2011blind,sun2013edge,mao2022deep}. Extensions to consider the non-uniform motion in real-world scenarios can be achieved by predicting pixel/patch-wise motion flows, where the optical flow should be estimated either from a single blurry image~\cite{gong2017motion} or consecutive frames~\cite{zhang2015intra}. Due to the development of deep neural networks, learning-based approaches have been widely applied for motion deblurring taking advantage of a large amount of data~\cite{gong2017motion,nimisha2017blur,jin2018learning,jin2019learning,purohit2019bringing, zhang2021exposure}. However, artificial and data-driven priors may suffer from inaccurate modeling of intra-frame textures or motions in the real world, preventing the generality of these algorithms.

\subsection{Stereo Motion Deblurring} 
An alternative direction of motion deblurring is the introduction of multiview images as a supplement \cite{sellent2016stereo,pan2019joint,zhou2019davanet}. 
With the stereo camera setup, images from two perspectives can be adopted to infer the scene depth, which strongly connects to the motion field~\cite{zhou2019davanet} and thus can be exploited to boost the deblurring performance. Sellent \etal, \cite{sellent2016stereo} generate a sharp, clear image from two stereo blurry frames. They induce blur kernels via local homographies and estimate optical flows to avoid ringing and boundary artifacts. DAVANet \cite{zhou2019davanet} is proposed to handle parallax by bidirectional disparities and varying information from two views to obtain stereo-sharp image pairs. Pan \etal, \cite{pan2019joint} propose a joint optimization framework to restore the latent images for generic dynamic scenes that benefited from incorporating 3D scene cues with pre-estimated scene flow and the improved boundaries information. However, the lack of temporal information prevents the above methods from generating sequences of sharp images.

\subsection{Event-based Motion Deblurring} 
Benefiting from the extremely low latency, events can provide the missing intra-frame information about motions and intensity textures~\cite{gallego2020event}, leading to event-based motion deblurring approaches. Wang \etal, \cite{wang2020event} propose eSL-Net to predict sharp images by integrating the events and frames into a sparsity framework. Still, eSL-Net cannot maintain its performance on real-world datasets due to its training on synthetic events. To bridge the gap between synthetic and real-world events, LEDVDI trains directly on real-world events \cite{lin2020learning}, while RED-Net exploits the blurry consistency and photometric consistency to enable semi-supervision on the deblurring network with both synthetic and real-world data \cite{xu2021motion}. Armed with supplementary events, event-based approaches always perform better than intensity-based deblurring approaches. However, existing approaches heavily rely on the assumption of the strict spatial alignment between the input blurry images and the event streams. They thus can not be easily employed in the general real-world scenarios, where the misalignment might exist, \eg, the \textit{DSEC} dataset based on the stereo event and intensity camera setup~\cite{gehrig2021dsec}. In this situation, the parallax between the event and intensity camera brings new challenges to the task of event-based motion deblurring. 

\subsection{Disparity Estimation}
Standard disparity estimation methods, with dual-intensity stereo cameras, employ neural networks to search the pixel-to-pixel correspondence between two views of the epipolar line, relying on the assumption that the stereo image pair shares the same modalities and ideal exposures \cite{shen2021cfnet,xu2022attention,li2022practical,xu2020aanet,li2021revisiting,zhang2021farther}. 
To reduce the dependence of the disparity estimation performance on high-quality imaging conditions, recent works replace one intensity camera with an event camera to build the stereo event and intensity camera setup and explore the cross-modal stereo matching task \cite{wang2021stereo, kim2022real, zuo2021accurate, gu2022self}. Due to the fact that event and intensity cameras perceive the same light field, the edge information extracted from events and intensity images is correlated to calculate the sparse disparity map \cite{wang2021stereo, kim2022real}. Meanwhile, several deep learning-based methods \cite{zuo2021accurate, gu2022self} are proposed to obtain dense results. Specifically, Zou \etal, \cite{zuo2021accurate} introduces the pyramid attention module to selectively combine the information of intensity images and asynchronous events. Gu \etal, \cite{gu2022self} propose a self-supervised learning framework, extending the same-modal cross-view warp consistency to cross-modal gradient structural consistency. However, the above methods fail to fully exploit the effectiveness of the stereo event and intensity camera setup in highly dynamic scenes, which motivates us to develop a new approach for unifying motion deblurring and disparity estimation.

Learning with the stereo event and intensity camera setup is more adaptive to highly dynamic scenes than the standard dual-intensity camera setup. Moreover, such a setup can benefit from the abundant motion and texture information due to the high temporal resolution of the event camera, which motivates us to develop a new model to learn event-based motion deblurring by spatially aligning real-world events and motion-blurred images.

\begin{figure*}[t]
    \centering
    \includegraphics[width=\textwidth]{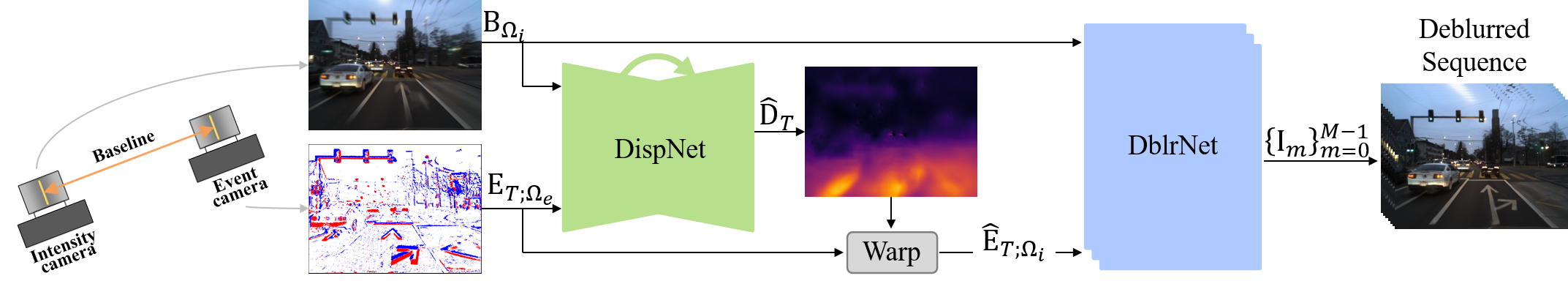}
    \caption{The overall structure of the proposed \myname, which consists of two modules: DispNet and DblrNet. DispNet estimates the coarse disparity between the blurry image and the corresponding event streams to realize the cross-modal pre-alignment of the inputs from two views. Then DblrNet receives the coarsely aligned data and generates a sequence of sharp and clear deblurring results.}
    \label{fig:pipline}
    \vspace{-2mm}
\end{figure*}

\section{Method}

\subsection{Problem Formulation}
\label{sec:problem}
Events, embedded with implicit information and textures, provide important assistance in motion deblurring, which introduces the task of Event-based motion Deblurring (ED). It focuses on reconstructing a sequence of sharp and clear latent images $\{\mathbf{I}_t\}_{{t}\in{T}}$ from a single motion blurred image $\mathbf{B}$ captured with exposure time $T$ and the corresponding events $\mathbf{E}_{T}$ triggered during $T$. Thus, the problem of ED can be formulated as,

\begin{equation}
    \mathbf{I}_{t;\Omega}=\text{ED}(\mathbf{B}_\Omega,\mathbf{E}_{T;\Omega}),\quad t\in{T}
    \label{eq:ed}
\end{equation}
where $\Omega$ indicates the same spatial domain shared by the intensity image and the corresponding event streams, which acts as the prior assumption and guides most ED methods to restore the latent sharp images from pixel-to-pixel aligned frames and event streams \cite{lin2020learning,wang2020event,xu2021motion}. However, it is not always correct in the real world due to the parallax between the event and intensity views, resulting in serious artifacts, as shown in \cref{fig:first}.

Motivated by the above problem, we propose a novel task of Event-based motion Deblurring with the STereo event and intensity cameras (St-ED), aiming at ED directly from the spatially misaligned inputs, \ie,

\begin{equation}
    \mathbf{I}_{t;\Omega_i}=\text{St-ED}(\mathbf{B}_{\Omega_i},\mathbf{E}_{T;\Omega_e}|\mathcal{M}_{\mathbf{D}_t}),\quad t\in{T}
    \label{eq:sted}
\end{equation}
where $\Omega_i$ / $\Omega_e$ respectively denote the spatial domain of the intensity image/events and $\mathcal{M}_{\mathbf{D}_t}:\Omega_e \mapsto \Omega_i$ denotes the pixel-level correspondence mapping operator parameterized by,
\begin{equation}
    \mathbf{D}_t\triangleq f(\mathbf{I}_{t;\Omega_i},\mathbf{E}_{T;\Omega_e})
    \label{eq:d_ie}
\end{equation}
where $\mathbf{D}_t$ indicates the disparities between the sharp latent image and event streams at the timestamp $t$ and $f(\cdot)$, denoting the disparity estimation function.

Ideally, we can consider St-ED as a two-step task by aligning with $\mathbf{D}_t$ and then deblurring following \cref{eq:ed}. However, to efficiently realize St-ED in real-world scenarios, challenges still exist.

\begin{itemize}
    \item The information loss of $\mathbf{B}_{\Omega_i}$ will bring mismatching and lead to incorrect mapping operator $\mathcal{M}_{\mathbf{D}_t}$. Meanwhile, it is necessary to align the blurry image and events for the deblurring process, \ie, an accurate mapping $\mathbf{D}_t$ is required. The above-coupled problem motivates us to consider a degenerate strategy for St-ED.
    \item Although the stereo event and intensity camera setup can directly obtain the ground-truth depths by introducing additional sensors such as LiDAR, its low frame rate and sparsity limits its perception of dynamic scene depths.
\end{itemize}

Therefore, we design \myname\ as a coarse-to-fine framework to first coarsely alleviate parallax between intensity and event views and then realize the fine fusion and enhancement for cross-modal data to reconstruct sharp images. 



\subsection{Network Architecture}
\cref{fig:pipline} illustrates the overall architecture of our network. \myname\ receives the blurry image $\mathbf{B}_{\Omega_i}$ and the corresponding event streams $\mathbf{E}_{T;\Omega_e}$ as input, and outputs a sequence of sharp latent images $\{\mathbf{I}_m\}^{M-1}_{m=0}$ and a disparity map $\mathbf{\hat{D}}_T$. Our network has two main modules: DispNet for coarsely aligning the blurry image and the event streams, and DblrNet for fine stereo matching and fusion of the coarsely aligned data to produce deblurred images.

\begin{figure}[t]
    \centering
    \includegraphics[width=\linewidth]{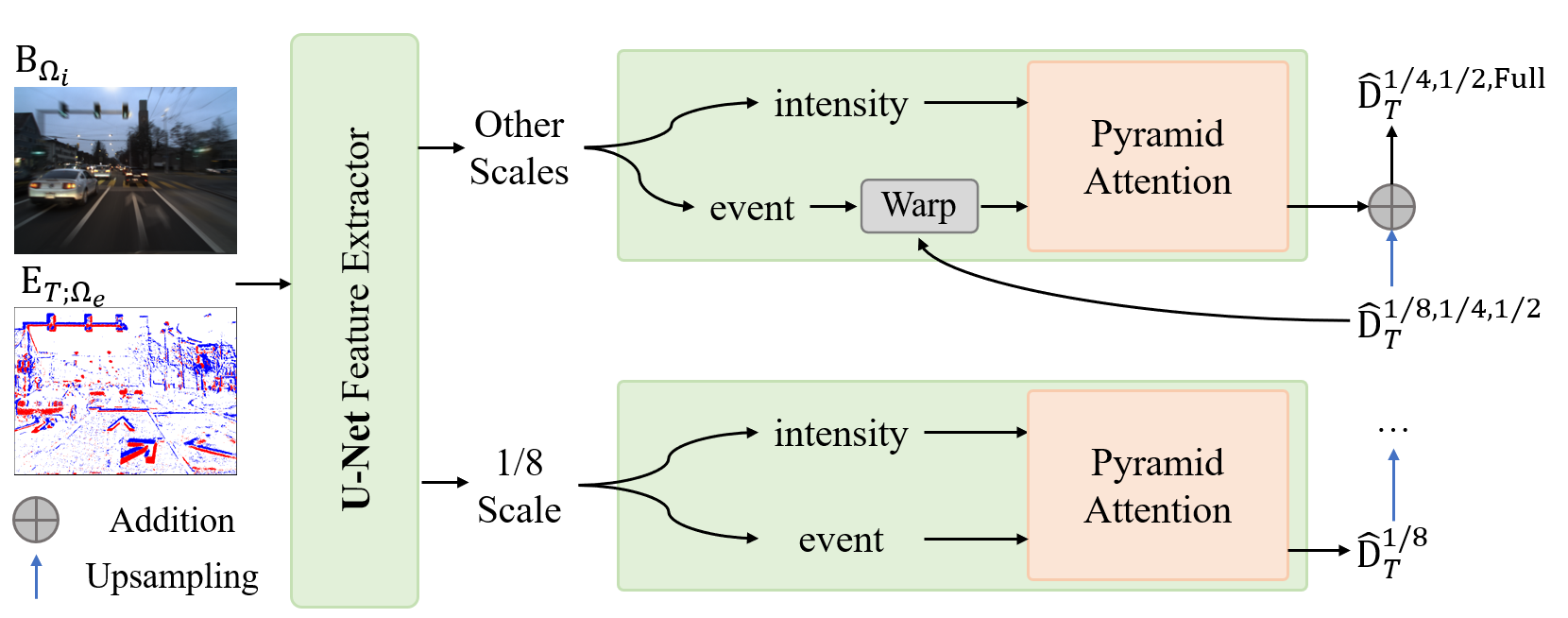}
    \caption{Overview of the DispNet.}
    \label{fig:dispnet}
    \vspace{-4mm}
\end{figure}

\begin{figure}
    \centering
    \includegraphics[width=\linewidth]{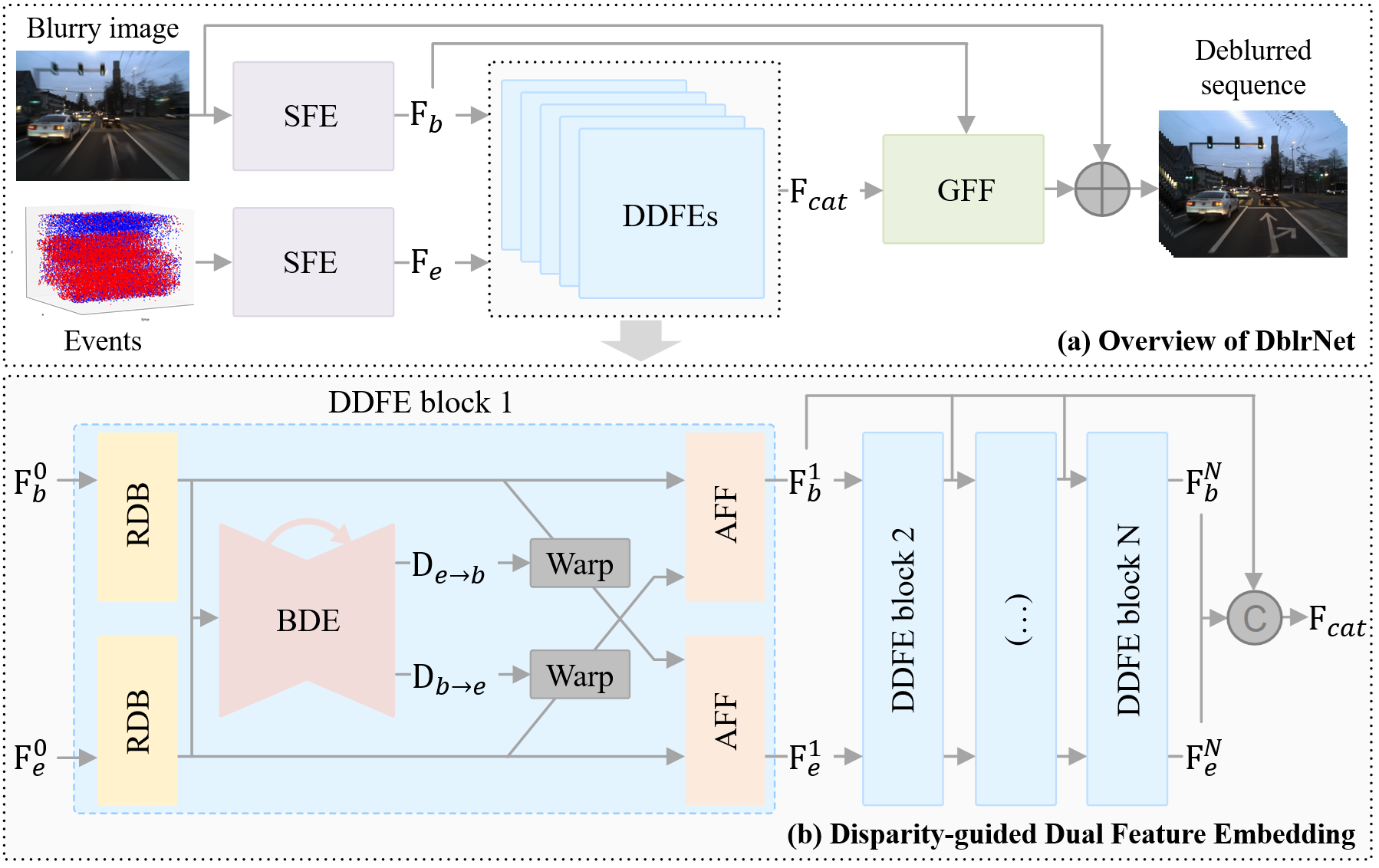}
    \caption{(a) Overview of the DblrNet. (b) Cascaded Disparity-guided Dual Feature Embedding (DDFE) module containing two Residual Dense Blocks (RDB), a Bidirectional Disparity Estimation (BDE) module, and two Attention-based Feature Fusion (AFF) blocks.}
    \label{fig:dblrnet}
    \vspace{-4mm}
\end{figure}

\subsubsection{DispNet}
Although, as mentioned in \cref{sec:problem}, unclear textures and edges in the blurry image prevent us from directly obtaining accurate disparity between two views in one step, we can still utilize the known information to approximate the coarse correspondence mapping from the spatial domain of the events to that of the image at the first step of St-ED. To this end, DispNet is designed to realize the degenerated version of \cref{eq:d_ie} as,
\begin{equation}
    \mathbf{\hat{D}}_T\approx f(\mathbf{B}_{\Omega_i},\mathbf{E}_{T;\Omega_e})
    \label{eq:d_be}
\end{equation}
roughly generalizing the correspondence between the blurry image and event streams during the exposure time $T$.

Compared with the original high-resolution blurry image, the low-resolution image produced by \textit{Down-sampling} or \textit{Pixel Unshuffle}\cite{shi2016real} has sharper edge and texture information \cite{gao2019dynamic, cho2021rethinking}. Hence, we adopt a U-Net-based architecture to estimate the coarse disparity between the blurry image and the event streams, which is fed with multi-modal data with various resolutions (1/8, 1/4, 1/2, 1) generated by \textit{Pixel Unshuffle}. As shown in \cref{fig:dispnet}, to ease the convergence of DispNet, we compute the initial disparity map at the first stage with the smallest resolution (1/8) and the residual maps at other stages (1/4, 1/2, and full scales) for subsequent corrections. Inspired by \cite{oktay2018attention}, Pyramid Attention (PA) blocks calculate attention weights at each stage to integrate features from two views, consisting of three parallel convolution layers with different scales, \ie, $1\times1$, $3\times3$, $5\times5$, in a pyramid structure \cite{li2018pyramid}.

The coarse disparity $\mathbf{\hat{D}}_T$ is then utilized to pre-align the blurry image and the events from two views by, 
\begin{equation}
    \mathbf{\hat{E}}_{T;\Omega_i}=\text{Warp}(\mathbf{E}_{T;\Omega_e}, \mathbf{\hat{D}}_T)
\end{equation}
where $\text{Warp}(\cdot)$ is a pixel-to-pixel spatial backward warping operator \cite{jaderberg2015spatial}. The visualization results in \cref{fig:abladisp} verify that DispNet does help to mitigate the artifacts caused by the misaligned data, \eg the zebra crossing.

\def\sswidth{0.33\linewidth} 

\begin{figure}[t]
    \begin{tabular}{c c c}
        \hspace{-2mm}\begin{tikzpicture}[spy using outlines={green,magnification=\ssmag,size=\ssizz},inner sep=0]
    		\node {\includegraphics[width=\sswidth]{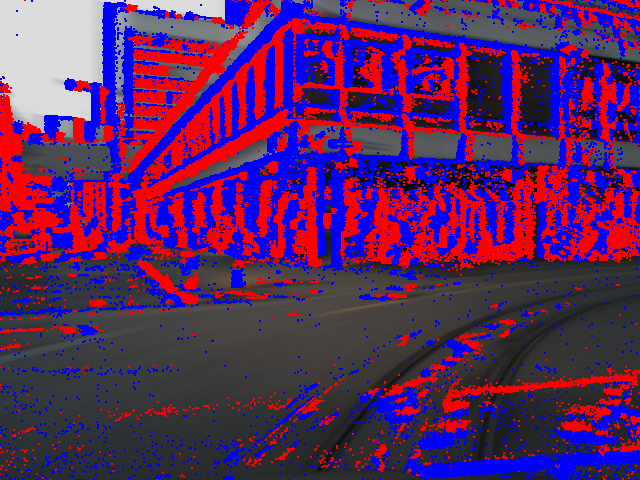}};
    	\end{tikzpicture}& \hspace{-4.8mm}
        \begin{tikzpicture}[spy using outlines={green,magnification=\ssmag,size=\ssizz},inner sep=0]
    		\node {\includegraphics[width=\sswidth]{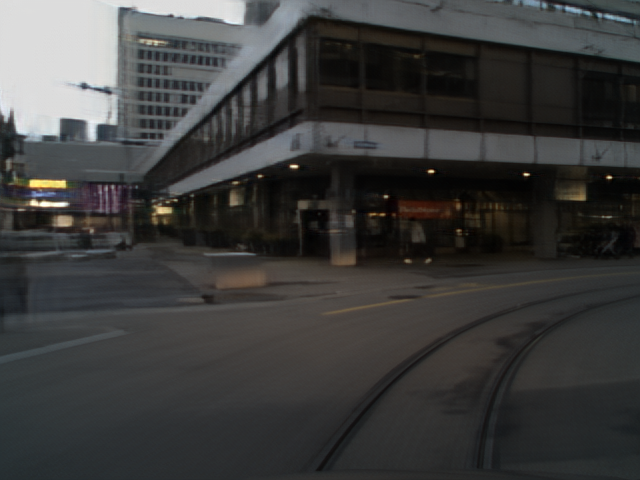}};
    	\end{tikzpicture}& \hspace{-4.8mm}
    	\begin{tikzpicture}[spy using outlines={green,magnification=\ssmag,size=\ssizz},inner sep=0]
    		\node {\includegraphics[width=\sswidth]{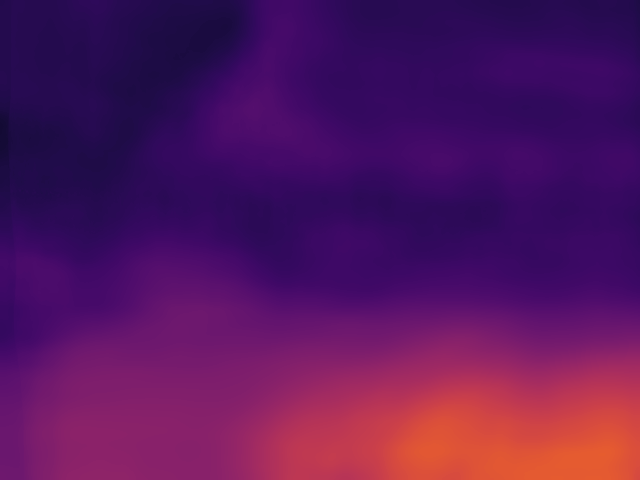}};
    	\end{tikzpicture}\vspace{-0.2mm}\\
    	\hspace{-2mm}\footnotesize{(a) Inputs} & \hspace{-2mm}\footnotesize{(b) Deblurred result} & \hspace{-2mm}\footnotesize{(c) $\mathbf{\hat{D}}_T$}\\
    	\hspace{-2mm}\begin{tikzpicture}[spy using outlines={green,magnification=\ssmag,size=\ssizz},inner sep=0]
    		\node {\includegraphics[width=\sswidth]{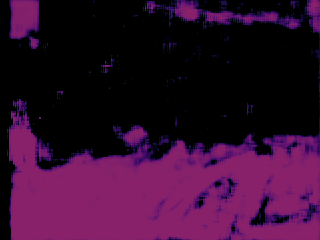}};
    	\end{tikzpicture}& \hspace{-4.8mm}
        \begin{tikzpicture}[spy using outlines={green,magnification=\ssmag,size=\ssizz},inner sep=0]
    		\node {\includegraphics[width=\sswidth]{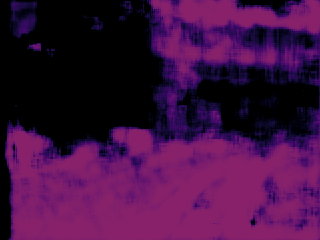}};
    	\end{tikzpicture}& \hspace{-4.8mm}
        \begin{tikzpicture}[spy using outlines={green,magnification=\ssmag,size=\ssizz},inner sep=0]
    		\node {\includegraphics[width=\sswidth]{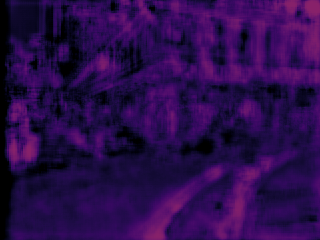}};
    	\end{tikzpicture}\vspace{-0.8mm}\\
        \hspace{-2mm}\footnotesize{(d) $\textbf{D}^{0}_{{b}\to{e};2}$} & \hspace{-2mm}\footnotesize{(e) $\textbf{D}^{1}_{{b}\to{e};2}$} & \hspace{-2mm}\footnotesize{(f) $\textbf{D}^{2}_{{b}\to{e};2}$}\\
    	\hspace{-2mm}\begin{tikzpicture}[spy using outlines={green,magnification=\ssmag,size=\ssizz},inner sep=0]
    		\node {\includegraphics[width=\sswidth]{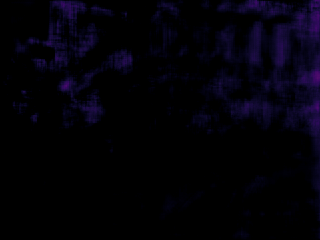}};
    	\end{tikzpicture}& \hspace{-4.8mm}
    	\begin{tikzpicture}[spy using outlines={green,magnification=\ssmag,size=\ssizz},inner sep=0]
    		\node {\includegraphics[width=\sswidth]{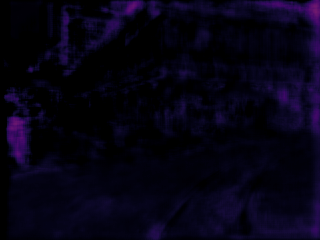}};
    	\end{tikzpicture}& \hspace{-4.8mm}
        \begin{tikzpicture}[spy using outlines={green,magnification=\ssmag,size=\ssizz},inner sep=0]
    		\node {\includegraphics[width=\sswidth]{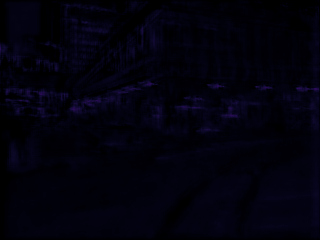}};
    	\end{tikzpicture}\vspace{-0.2mm}\\
    	\hspace{-2mm}\footnotesize{(g) $\textbf{D}^{3}_{{b}\to{e};2}$} & \hspace{-2mm}\footnotesize{(h) $\textbf{D}^{4}_{{b}\to{e};2}$} & \hspace{-2mm}\footnotesize{(i) $\textbf{D}^{5}_{{b}\to{e};2}$}\\
    \end{tabular}
    \caption{Intermediate disparities predicted by (c) DispNet, and (d-i) BDE modules in DeblurNet. We only show the $l=2^{nd}$ patch disparities of each BDE.}
    \label{fig:bde}
\end{figure}

\subsubsection{DblrNet} 
The residual network architecture with dense connection is used as the backbone of our DblrNet, as shown in \cref{fig:dblrnet}. It first separately initializes the input blurry image $\mathbf{B}_{\Omega_i}$, and the coarsely aligned event streams $\mathbf{\hat{E}}_{T;\Omega_i}$ by Shallow Feature Extraction (SFE) modules into feature maps with $C$ channels, denoted by $\mathbf{F}^0_b, \mathbf{F}^0_e\in\mathbb{R}^{C\times\frac{H}{2}\times\frac{W}{2}}$. Then, the dual-path features are fed into $N$ cascaded Disparity-guided Dual Feature Embedding (DDFE) modules, gradually building the fine correspondence between the intensity and event views for effective feature alignment and fusion from dual sources. A Global Feature Fusion (GFF) module receives the learned high-dimensional feature $\mathbf{F}_{cat}$, assembling by concatenating the output blurry-path features of each DDFE module $\{\mathbf{F}^{i}_b\}^{N}_{i=1}$ and the output event-path features of the final DDFE module $\mathbf{F}^{N}_e$, and finally reconstructs a sequence of sharp images.

The major difference of our approach from the traditional event-based motion deblurring methods \cite{lin2020learning, xu2021motion, song2022cir} is that the latter directly concatenates the blurry image and the corresponding event streams as the initial fusion, acting as the input fed into the networks. This approach does not work when the precondition of the pixel-to-pixel spatial alignment does not hold in the stereo event and intensity camera setup. By contrast, our approach encodes and learns the multi-view data with two separate paths and estimates fine bidirectional disparities in high-dimensional feature levels to align and fuse the information from two paths.

\begin{figure}
	\centering
    \begin{minipage}[b]{\linewidth}
	    \centering
	    \subfloat[Stereo hybrid camera system.]{\includegraphics[width=0.75\linewidth]{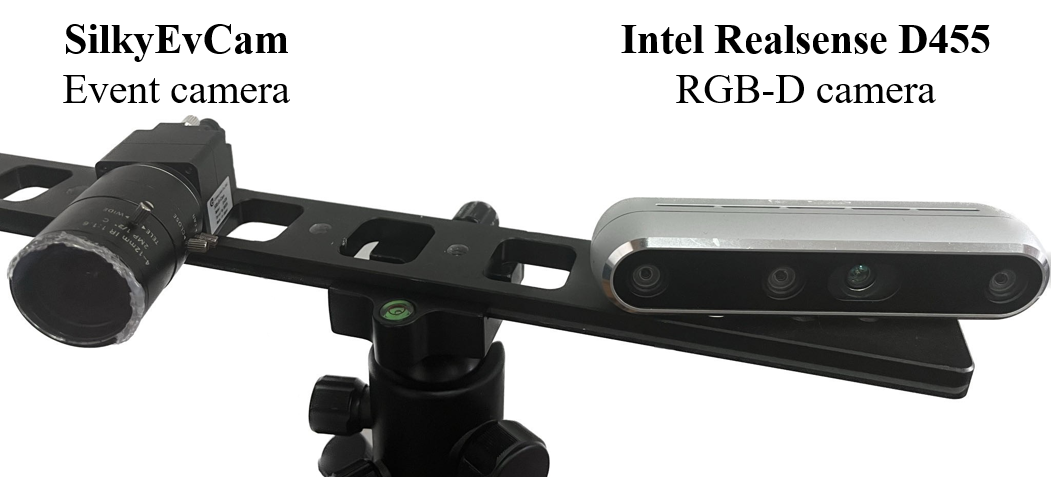}\label{fig:camera_cam}}
    \end{minipage}\\
    \begin{minipage}[b]{\linewidth}
	    \centering
	    \subfloat[Examples of scenes in the proposed \textit{\mydata} dataset.]{\includegraphics[width=\linewidth]{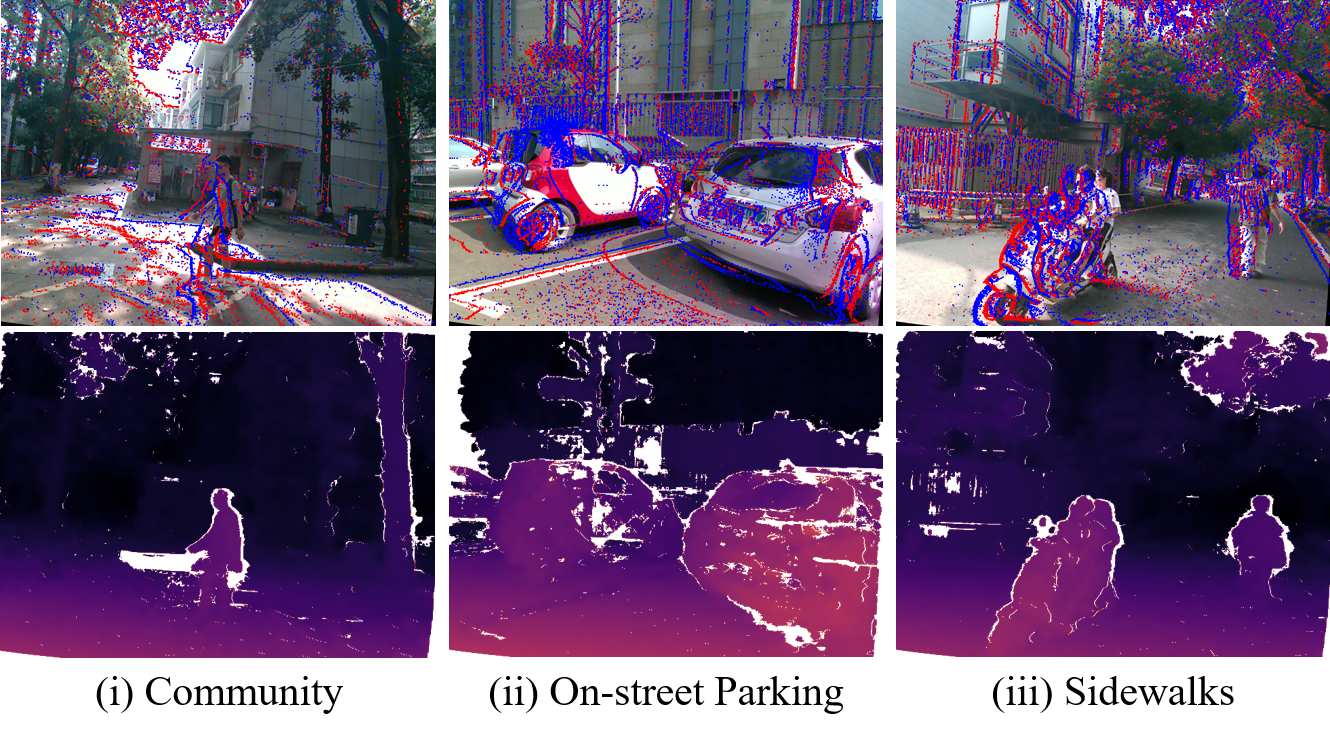}\label{fig:camera_scene}}
    \end{minipage}
	\caption{(a) Illustration of the stereo event and intensity camera setup. It features a SilkyEvCam event camera (left) and an Intel Realsense D455 RGB-D camera (right). (b) Example images rendered with events (top) and the corresponding disparity images (bottom).}
	\label{fig:camera}
	\vspace{-4mm}
\end{figure}

The Bidirectional Disparity Estimation (BDE) module and two following backward warping operators are the core architectures of DDFE. Specifically, in the $i$-th $(1\leqslant i \leqslant N)$ DDFE module, BDE receives dual-path features $\mathbf{F}^{i-1}_b$ and $\mathbf{F}^{i-1}_e$ that are updated by two parallel Residual Dense Blocks (RDB). For the reason that the multi-head warping enhances the expressiveness of the feature descriptor by splitting the channel dimension of the feature map into $L$ groups, BDE estimates bidirectional disparities $\textbf{D}^{i-1}_{{b}\to{e}}$ and $\textbf{D}^{i-1}_{{e}\to{b}}$ by,
\begin{equation}
    [\textbf{D}^{i-1}_{{b}\to{e}};\textbf{D}^{i-1}_{{e}\to{b}}]=\text{BDE}(\mathbf{F}^{i-1}_\mathbf{B}, \mathbf{F}^{i-1}_\mathbf{E}),
    \label{eq:mse}
\end{equation}
with $\textbf{D}^{i-1}_{{b}\to{e}}=\{\textbf{D}^{i-1}_{{b}\to{e};l}\}^{L-1}_{l=0}$ and $\textbf{D}^{i-1}_{{e}\to{b}}=\{\textbf{D}^{i-1}_{{e}\to{b};l}\}^{L-1}_{l=0}$.
The intermediate disparities in \cref{fig:bde} show that the value of the bidirectional disparities decreases as the number of DDFE modules increases, demonstrating that our DblrNet can gradually align and fuse dual feature maps from intensity and event views. Similarly, we split the channel dimension of the feature map $\textbf{F}$ into $L$ groups, denoted by ${\mathbf{F}_{0},\dots,\mathbf{F}_{L-1}}\in\mathbb{R}^{\frac{C}{L}\times\frac{H}{2}\times\frac{W}{2}}$. $\mathbf{D}_{{b}\to{e}}$ and $\mathbf{D}_{{e}\to{b}}$ are then used to warp each patch of feature map $\mathbf{F}_l$ from one path to the other by channel. In our implementation, the number $L$ is set to 6. For the $l$-th patch, the above process can be presented by,
\begin{equation}
    \begin{aligned}
        \mathbf{W}^{i-1}_{b;l}=\text{Warp}(\mathbf{F}^{i-1}_{b;l},\mathbf{D}^{i-1}_{{e}\to{b};l})\\
        \mathbf{W}^{i-1}_{e;l}=\text{Warp}(\mathbf{F}^{i-1}_{e;l},\mathbf{D}^{i-1}_{{b}\to{e};l})
    \end{aligned}, \quad l=0,\dots,L-1.
    \label{eq:warp1}
\end{equation}

\begin{table}[t]
    \centering
    \caption{Comparisons of our \textit{\mydata} with two publicly available datasets, \ie, \textit{MVSEC} \cite{zhu2018multivehicle} and \textit{DSEC} \cite{gehrig2021dsec}.}
    \begin{tabular}{l c p{0.5cm}<{\centering} p{0.8cm}<{\centering} p{0.8cm}<{\centering} c}
    \toprule[1.pt]
    \multirow{2}{*}{Dataset} & \multirow{2}{*}{Resolution} & \multirow{2}{*}{Color} & FPS & FPS & Density of  \\ 
    & & & (image) & (depth) & Depth \\
    \hline
    MVSEC \cite{zhu2018multivehicle} & $346\times 260$ & \ding{56} & 32 & 20 & Sparse \\
    DSEC \cite{gehrig2021dsec} & $\mathbf{640} \times \mathbf{480}$ & \ding{52} & 20 & 10 & Sparse \\ 
    \hline
    StEIC (Ours) & $\mathbf{640} \times \mathbf{480}$ & \ding{52} & \textbf{60} & \textbf{60} & \textbf{Dense} \\ 
    \toprule[1.pt]
    \end{tabular}
    \label{tab:dataset}
\end{table}

\begin{table}[t]
    \centering
    \caption{Details about the proposed \textit{\mydata} dataset. In total, the dataset consists of 65 sequences for real-world street scenarios.}
    \begin{tabular}{llccc}
    \toprule[1.pt] 
    Split & Area & \#Seq & \#Pairs & \#Events (M)\\
    \hline
    \textbf{Train} & Community & 15 & 1037 & 1453 \\
    & On-street Parking & 14 & 1628 & 1634 \\
    & Sidewalks & 14 & 1037 & 1005 \\
    \cline{3-5} 
    & & \textbf{43} & \textbf{3702} & \textbf{4092} \\ 
    \hline
    \textbf{Test} & Community & 9 & 631 & 687 \\
    & On-street Parking & 6 & 788 & 464 \\
    & Sidewalks & 7 & 540 & 773 \\
    \cline{3-5} 
    & & \textbf{22} & \textbf{1959} & \textbf{1924} \\ 
    \hline
    \textbf{Total} & & \textbf{65} & \textbf{5661} & \textbf{6016}\\ 
    \toprule[1.pt]
    \end{tabular}
    \vspace{-1em}
    \label{tab:dataset2}
\end{table}

For two-view aggregation, the Attention-based Feature Fusion (AFF) module is exploited for each patch to actively emphasize or suppress the features from the self-path and the warped features from the other path by,
\begin{equation}
    \begin{aligned}
        \mathbf{F}^{i}_{b}=\text{AFF}(\mathbf{F}^{i-1}_{b},\mathbf{W}^{i-1}_{e}) \\
        \mathbf{F}^{i}_{e}=\text{AFF}(\mathbf{F}^{i-1}_{e},\mathbf{W}^{i-1}_{b})
    \end{aligned}.
\end{equation}

Note that the BDE and AFF modules of each DDFE share their weights to reduce occupied memory.

\subsection{Losses}
\label{sec:loss}
Because intensity images and events share strong structural correlations, and \myname\ utilizes the $\text{Warp}(\cdot)$ operator to achieve disparity-based alignment of two-view data, \myname\ can produce a sequence of deblurred frames $\{\mathbf{I}_m\}^{M-1}_{m=0}$ and a disparity map $\mathbf{\hat{D}}_T$, only requiring sharp images as supervision without introducing ground-truth disparity.
Our training loss comprises three parts: deblurring loss $\mathcal{L}_{dblr}$, perceptual loss $\mathcal{L}_{perc}$, and disparity smoothness loss $\mathcal{L}_{tv}$.

\subsubsection{Deblurring Loss} We use the sequence of the sharp images $\{\mathbf{G}_m\}^{M-1}_{m=0}$ as ground-truth signals to supervise the network via the $\ell_1$ loss function:
\begin{equation}
    \mathcal{L}_{dblr} = \frac{1}{M}\sum^{M-1}_{m=0}\|\mathbf{I}_m-\mathbf{G}_m\|_1.
\end{equation}

\subsubsection{Perceptual Loss} The second loss function is the perceptual loss proposed in \cite{johnson2016perceptual}, which is defined as the $\ell_2$-norm between the VGG-19 features of the deblurred frames and clear images:
\begin{equation}
    \mathcal{L}_{perc} = \frac{1}{M}\sum^{M-1}_{m=0}\|\Phi_j(\mathbf{I}_m)-\Phi_j(\mathbf{G}_m)\|^2_2,
\end{equation}
where $\Phi_j(\cdot)$ denotes the features from the $j$-th convolution layer within the pre-trained VGG-19 network. In our work, we use the features from the conv3-3 layer ($j=15$).

In addition to the two losses mentioned above, we also employ the Total Variation (TV) loss to smooth the disparity maps predicted in \myname. The overall loss function $\mathcal{L}$ can be defined as
\begin{equation}
    \mathcal{L} = \lambda_1\mathcal{L}_{dblr} + \lambda_2\mathcal{L}_{perc} + \lambda_3\mathcal{L}_{tv},
\end{equation}
with $\lambda_1$, $\lambda_2$ and $\lambda_3$ denoting the balancing parameters.

\def\ssxxsone{(0.15,-0.25)} 
\def\ssxxsfour{(-0.5,0.75)} 
\def\ssyysone{(0.54,-0.68)} 
\def\ssxxstwo{(-1,1.45)} 
\def\ssxxsthree{(0.1,0.85)} 
\def\ssyystwo{(-0.54,-0.68)} 
\def\ssizz{1.15cm} 
\def\sswidth{0.1975\textwidth} 
\def\ssmagtwo{3.5}
\def\ssmag{4}
\def\scc{(-1.2,0.92)}
\def\sccone{(-1.75,1.175)}

\begin{figure*}[t]
	\centering
	\begin{tikzpicture}[spy using outlines={yellow,magnification=\ssmag,size=\ssizz},inner sep=0]
		\node {\includegraphics[width=\sswidth]{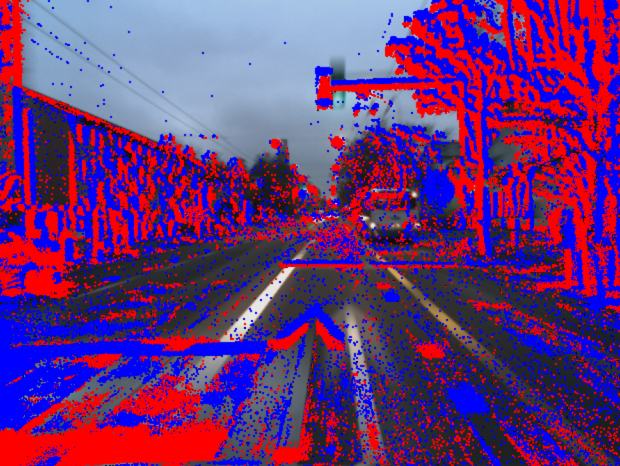}};
		\spy on \ssxxsthree in node [left] at \ssyystwo;
        \node [anchor=west] at \sccone {\textcolor{lime}{\bf Inputs}};
	\end{tikzpicture} \hspace{-2.2mm}
	\begin{tikzpicture}[spy using outlines={yellow,magnification=\ssmag,size=\ssizz},inner sep=0]
		\node {\includegraphics[width=\sswidth]{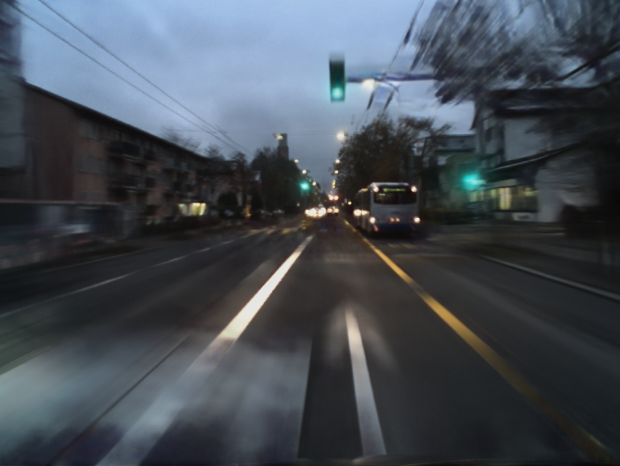}};
		\spy on \ssxxsthree in node [left] at \ssyystwo;
		\node [anchor=west] at \sccone {\textcolor{lime}{\bf DAVANet}};
	\end{tikzpicture} \hspace{-2.2mm}
	\begin{tikzpicture}[spy using outlines={yellow,magnification=\ssmag,size=\ssizz},inner sep=0]
		\node {\includegraphics[width=\sswidth]{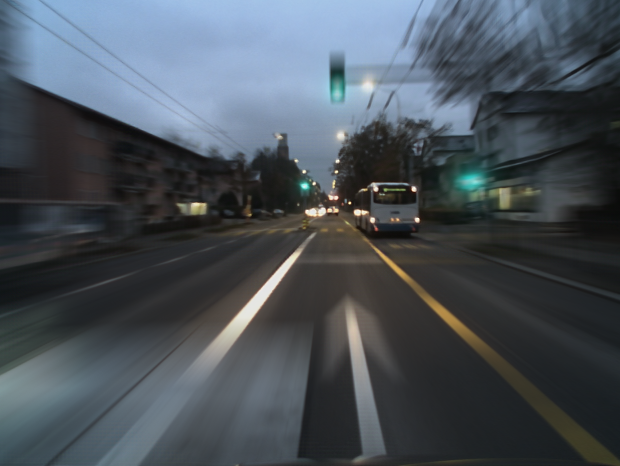}};
		\spy on \ssxxsthree in node [left] at \ssyystwo;
		\node [anchor=west] at \sccone {\textcolor{lime}{\bf LEVS}};
	\end{tikzpicture} \hspace{-2.2mm}
	\begin{tikzpicture}[spy using outlines={yellow,magnification=\ssmag,size=\ssizz},inner sep=0]
		\node {\includegraphics[width=\sswidth]{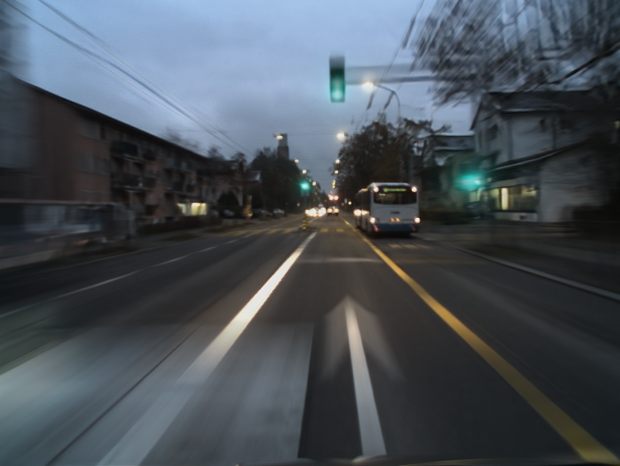}};
		\spy on \ssxxsthree in node [left] at \ssyystwo;
		\node [anchor=west] at \sccone {\textcolor{lime}{\bf Motion-ETR}};
	\end{tikzpicture} \hspace{-2.2mm}
	\begin{tikzpicture}[spy using outlines={yellow,magnification=\ssmag,size=\ssizz},inner sep=0]
		\node {\includegraphics[width=\sswidth]{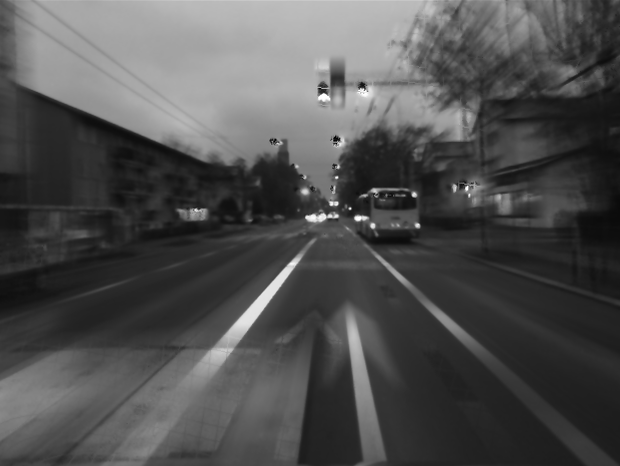}};
		\spy on \ssxxsthree in node [left] at \ssyystwo;
		\node [anchor=west] at \sccone {\textcolor{lime}{\bf EDI}};
	\end{tikzpicture} \vspace{0.25mm} \\

	\begin{tikzpicture}[spy using outlines={yellow,magnification=\ssmag,size=\ssizz},inner sep=0]
		\node {\includegraphics[width=\sswidth]{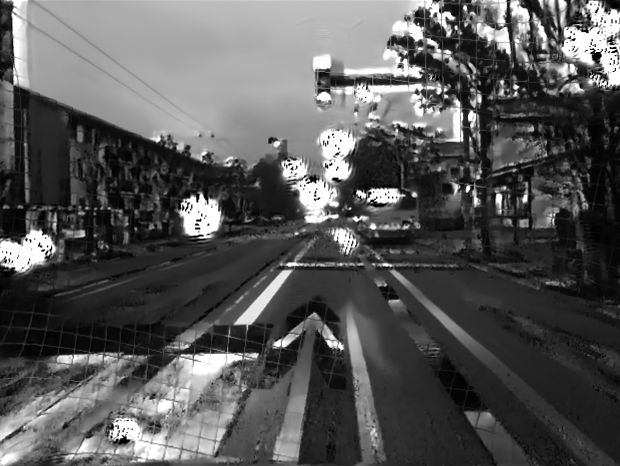}};
		\spy on \ssxxsthree in node [left] at \ssyystwo;
		\node [anchor=west] at \sccone {\textcolor{lime}{\bf eSL-Net}};
	\end{tikzpicture} \hspace{-2.2mm}
	\begin{tikzpicture}[spy using outlines={yellow,magnification=\ssmag,size=\ssizz},inner sep=0]
		\node {\includegraphics[width=\sswidth]{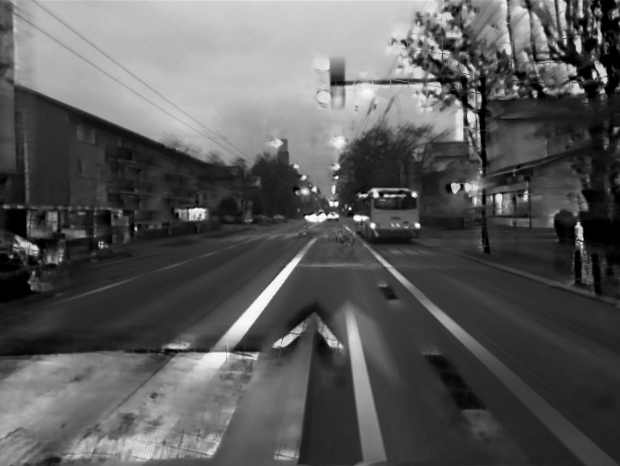}};
		\spy on \ssxxsthree in node [left] at \ssyystwo;
		\node [anchor=west] at \sccone {\textcolor{lime}{\bf LEDVDI}};
	\end{tikzpicture} \hspace{-2.2mm}
	\begin{tikzpicture}[spy using outlines={yellow,magnification=\ssmag,size=\ssizz},inner sep=0]
		\node {\includegraphics[width=\sswidth]{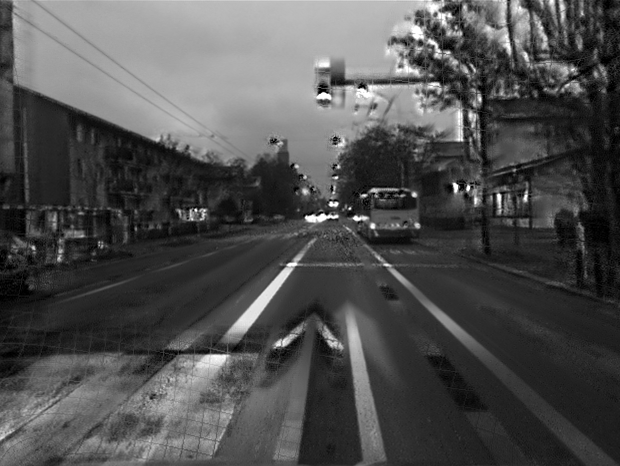}};
		\spy on \ssxxsthree in node [left] at \ssyystwo;
		\node [anchor=west] at \sccone {\textcolor{lime}{\bf RED-Net}};
	\end{tikzpicture} \hspace{-2.2mm}
	\begin{tikzpicture}[spy using outlines={yellow,magnification=\ssmag,size=\ssizz},inner sep=0]
		\node {\includegraphics[width=\sswidth]{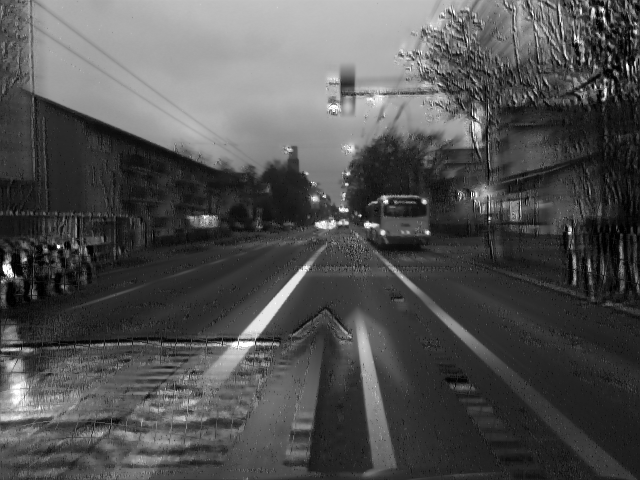}};
		\spy on \ssxxsthree in node [left] at \ssyystwo;
		\node [anchor=west] at \sccone {\textcolor{lime}{\bf E-CIR}};
	\end{tikzpicture} \hspace{-2.2mm}
	\begin{tikzpicture}[spy using outlines={yellow,magnification=\ssmag,size=\ssizz},inner sep=0]
		\node {\includegraphics[width=\sswidth]{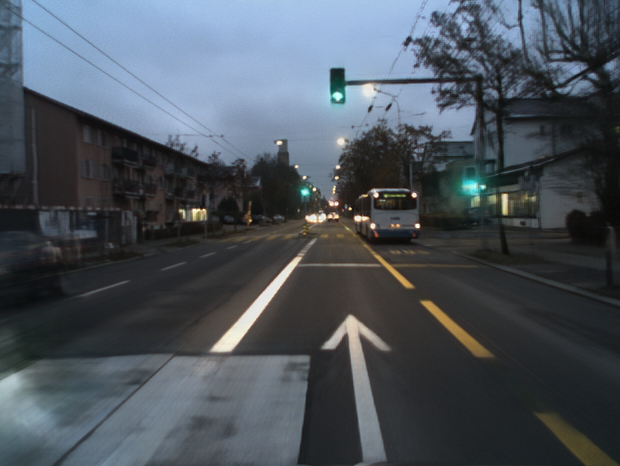}};
		\spy on \ssxxsthree in node [left] at \ssyystwo;
		\node [anchor=west] at \sccone {\textcolor{lime}{\bf \myname}};
	\end{tikzpicture} \vspace{0.5mm} \\ 
 
	\begin{tikzpicture}[spy using outlines={yellow,magnification=\ssmagtwo,size=\ssizz},inner sep=0]
		\node {\includegraphics[width=\sswidth]{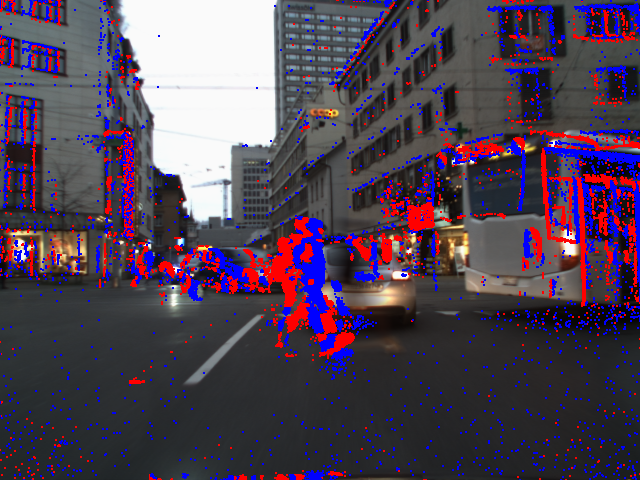}};
		\spy on \ssxxsone in node [left] at \ssyystwo;
		\node [anchor=west] at \sccone {\textcolor{lime}{\bf Inputs}};
	\end{tikzpicture} \hspace{-2.2mm}
	\begin{tikzpicture}[spy using outlines={yellow,magnification=\ssmagtwo,size=\ssizz},inner sep=0]
		\node {\includegraphics[width=\sswidth]{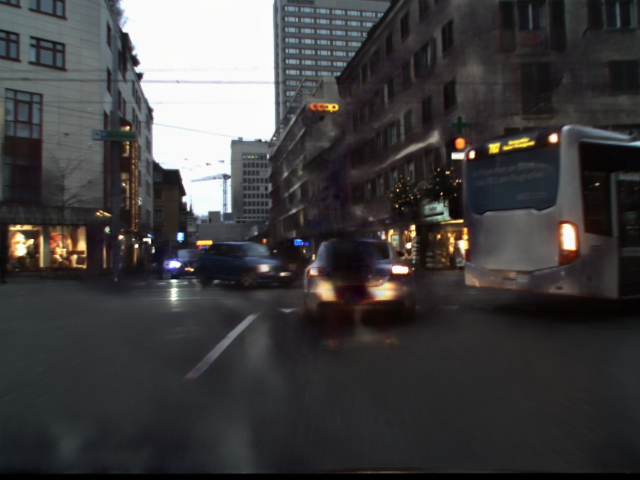}};
		\spy on \ssxxsone in node [left] at \ssyystwo;
		\node [anchor=west] at \sccone {\textcolor{lime}{\bf DAVANet}};
	\end{tikzpicture} \hspace{-2.2mm}
	\begin{tikzpicture}[spy using outlines={yellow,magnification=\ssmagtwo,size=\ssizz},inner sep=0]
		\node {\includegraphics[width=\sswidth]{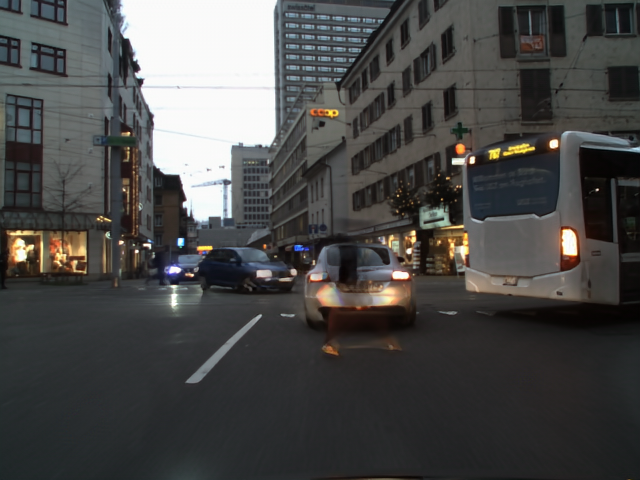}};
		\spy on \ssxxsone in node [left] at \ssyystwo;
		\node [anchor=west] at \sccone {\textcolor{lime}{\bf LEVS}};
	\end{tikzpicture} \hspace{-2.2mm}
	\begin{tikzpicture}[spy using outlines={yellow,magnification=\ssmagtwo,size=\ssizz},inner sep=0]
		\node {\includegraphics[width=\sswidth]{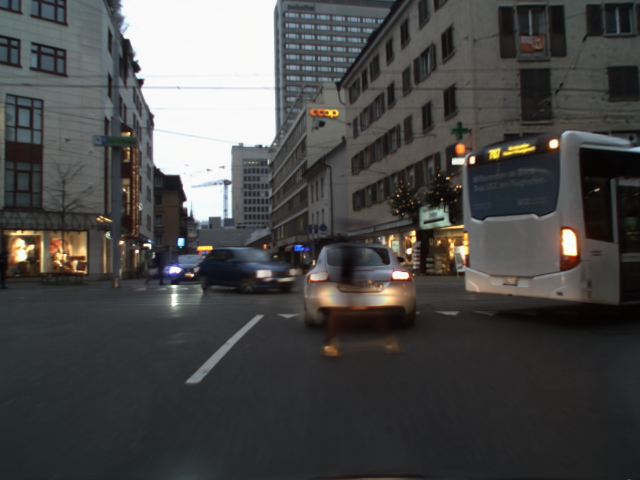}};
		\spy on \ssxxsone in node [left] at \ssyystwo;
		\node [anchor=west] at \sccone {\textcolor{lime}{\bf Motion-ETR}};
	\end{tikzpicture} \hspace{-2.2mm}
	\begin{tikzpicture}[spy using outlines={yellow,magnification=\ssmagtwo,size=\ssizz},inner sep=0]
		\node {\includegraphics[width=\sswidth]{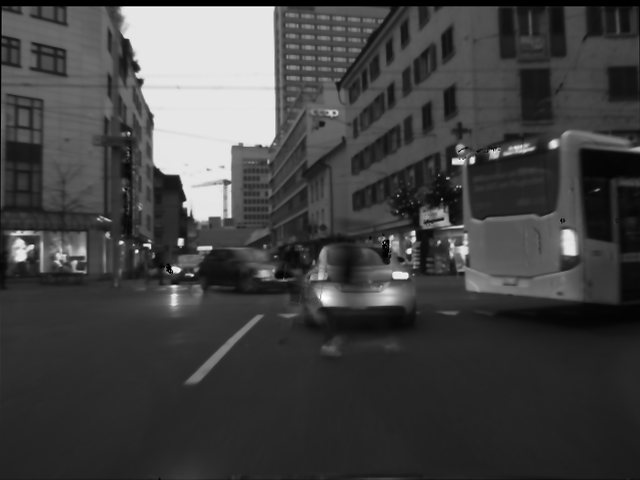}};
		\spy on \ssxxsone in node [left] at \ssyystwo;
		\node [anchor=west] at \sccone {\textcolor{lime}{\bf EDI}};
	\end{tikzpicture} \vspace{0.25mm} \\

	\begin{tikzpicture}[spy using outlines={yellow,magnification=\ssmagtwo,size=\ssizz},inner sep=0]
		\node {\includegraphics[width=\sswidth]{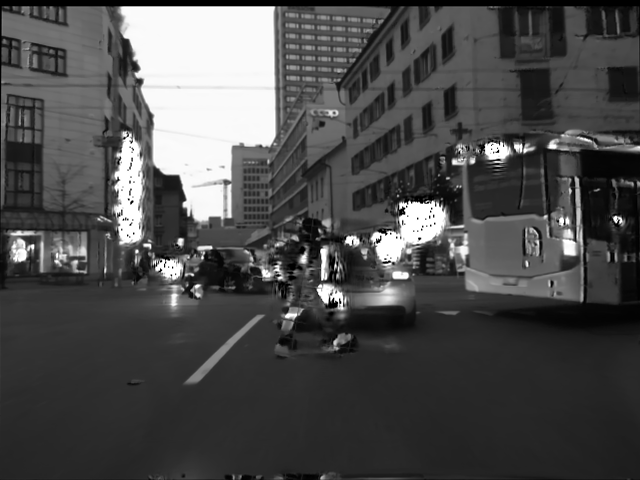}};
		\spy on \ssxxsone in node [left] at \ssyystwo;
		\node [anchor=west] at \sccone {\textcolor{lime}{\bf eSL-Net}};
	\end{tikzpicture} \hspace{-2.2mm}
	\begin{tikzpicture}[spy using outlines={yellow,magnification=\ssmagtwo,size=\ssizz},inner sep=0]
		\node {\includegraphics[width=\sswidth]{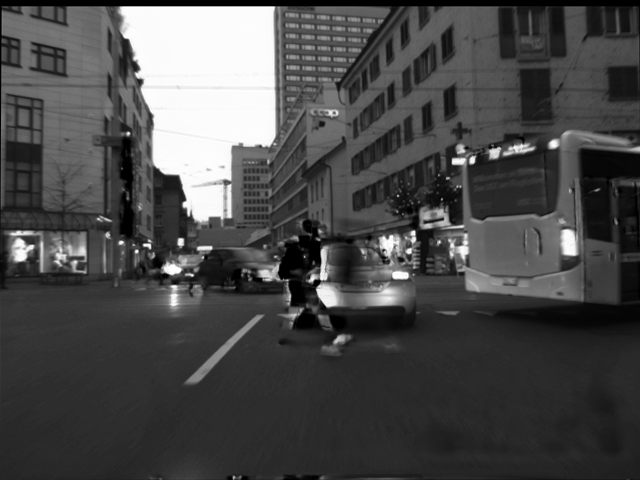}};
		\spy on \ssxxsone in node [left] at \ssyystwo;
		\node [anchor=west] at \sccone {\textcolor{lime}{\bf LEDVDI}};
	\end{tikzpicture} \hspace{-2.2mm}
	\begin{tikzpicture}[spy using outlines={yellow,magnification=\ssmagtwo,size=\ssizz},inner sep=0]
		\node {\includegraphics[width=\sswidth]{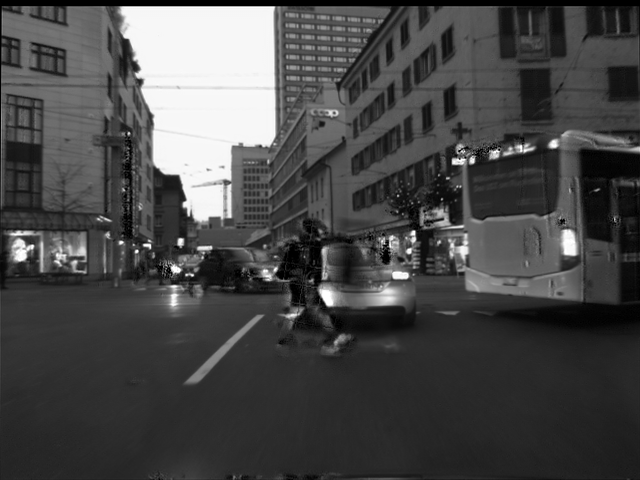}};
		\spy on \ssxxsone in node [left] at \ssyystwo;
		\node [anchor=west] at \sccone {\textcolor{lime}{\bf RED-Net}};
	\end{tikzpicture} \hspace{-2.2mm}
	\begin{tikzpicture}[spy using outlines={yellow,magnification=\ssmagtwo,size=\ssizz},inner sep=0]
		\node {\includegraphics[width=\sswidth]{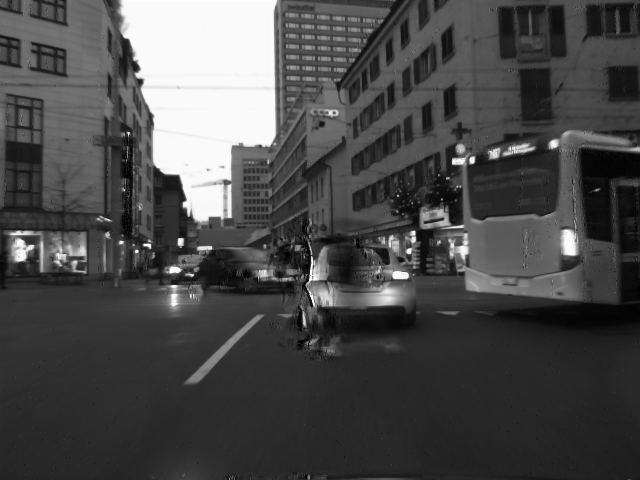}};
		\spy on \ssxxsone in node [left] at \ssyystwo;
		\node [anchor=west] at \sccone {\textcolor{lime}{\bf E-CIR}};
	\end{tikzpicture} \hspace{-2.2mm}
	\begin{tikzpicture}[spy using outlines={yellow,magnification=\ssmagtwo,size=\ssizz},inner sep=0]
		\node {\includegraphics[width=\sswidth]{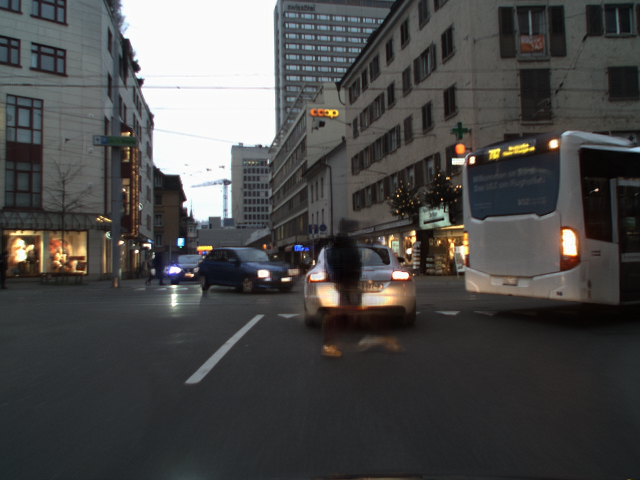}};
		\spy on \ssxxsone in node [left] at \ssyystwo;
		\node [anchor=west] at \sccone {\textcolor{lime}{\bf \myname}};
	\end{tikzpicture} \vspace{0.5mm} \\
	\caption{Qualitative results of motion deblurring of 9 different methods on the \textit{DSEC-large} dataset. We only select two exemplar frames for visualization.}
	\label{fig:qualitative_dsec}
	\vspace{-4mm}
\end{figure*}

\def\ssxxsone{(-0.05,0.75)} 
\def\ssxxsfour{(-0.5,0.75)} 
\def\ssyysone{(0.54,-0.68)} 
\def\ssxxstwo{(-1,1.45)} 
\def\ssxxsthree{(0.,0.85)} 
\def\ssyystwo{(-0.54,-0.68)} 
\def\ssizz{1.15cm} 
\def\sswidth{0.1975\textwidth} 
\def\ssmagtwo{4}
\def\ssmag{4}
\def\scc{(-1.2,0.92)}
\def\sccone{(-1.75,1.175)}

\begin{figure*}[t]
	\centering
	\begin{tikzpicture}[spy using outlines={yellow,magnification=\ssmag,size=\ssizz},inner sep=0]
		\node {\includegraphics[width=\sswidth]{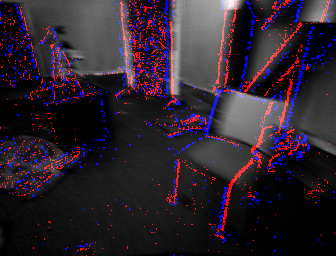}};
		\spy on \ssxxsthree in node [left] at \ssyystwo;
        \node [anchor=west] at \sccone {\textcolor{lime}{\bf Inputs}};
	\end{tikzpicture} \hspace{-2.2mm}
	\begin{tikzpicture}[spy using outlines={yellow,magnification=\ssmag,size=\ssizz},inner sep=0]
		\node {\includegraphics[width=\sswidth]{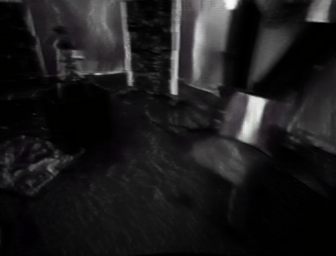}};
		\spy on \ssxxsthree in node [left] at \ssyystwo;
		\node [anchor=west] at \sccone {\textcolor{lime}{\bf DAVANet}};
	\end{tikzpicture} \hspace{-2.2mm}
	\begin{tikzpicture}[spy using outlines={yellow,magnification=\ssmag,size=\ssizz},inner sep=0]
		\node {\includegraphics[width=\sswidth]{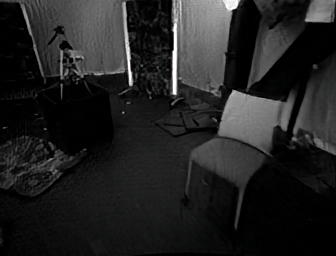}};
		\spy on \ssxxsthree in node [left] at \ssyystwo;
		\node [anchor=west] at \sccone {\textcolor{lime}{\bf LEVS}};
	\end{tikzpicture} \hspace{-2.2mm}
	\begin{tikzpicture}[spy using outlines={yellow,magnification=\ssmag,size=\ssizz},inner sep=0]
		\node {\includegraphics[width=\sswidth]{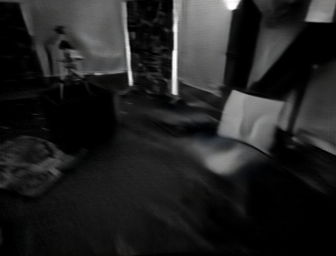}};
		\spy on \ssxxsthree in node [left] at \ssyystwo;
		\node [anchor=west] at \sccone {\textcolor{lime}{\bf Motion-ETR}};
	\end{tikzpicture} \hspace{-2.2mm}
	\begin{tikzpicture}[spy using outlines={yellow,magnification=\ssmag,size=\ssizz},inner sep=0]
		\node {\includegraphics[width=\sswidth]{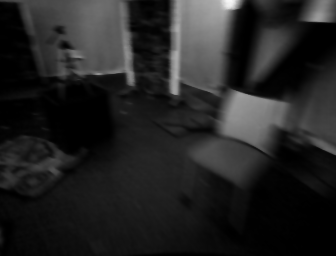}};
		\spy on \ssxxsthree in node [left] at \ssyystwo;
		\node [anchor=west] at \sccone {\textcolor{lime}{\bf EDI}};
	\end{tikzpicture} \vspace{0.25mm} \\

	\begin{tikzpicture}[spy using outlines={yellow,magnification=\ssmag,size=\ssizz},inner sep=0]
		\node {\includegraphics[width=\sswidth]{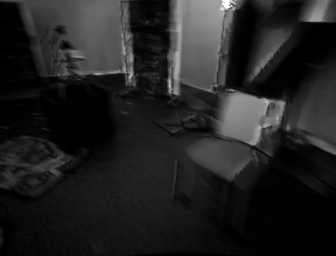}};
		\spy on \ssxxsthree in node [left] at \ssyystwo;
		\node [anchor=west] at \sccone {\textcolor{lime}{\bf eSL-Net}};
	\end{tikzpicture} \hspace{-2.2mm}
	\begin{tikzpicture}[spy using outlines={yellow,magnification=\ssmag,size=\ssizz},inner sep=0]
		\node {\includegraphics[width=\sswidth]{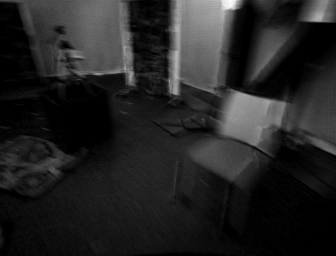}};
		\spy on \ssxxsthree in node [left] at \ssyystwo;
		\node [anchor=west] at \sccone {\textcolor{lime}{\bf LEDVDI}};
	\end{tikzpicture} \hspace{-2.2mm}
	\begin{tikzpicture}[spy using outlines={yellow,magnification=\ssmag,size=\ssizz},inner sep=0]
		\node {\includegraphics[width=\sswidth]{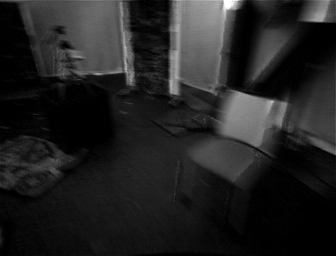}};
		\spy on \ssxxsthree in node [left] at \ssyystwo;
		\node [anchor=west] at \sccone {\textcolor{lime}{\bf RED-Net}};
	\end{tikzpicture} \hspace{-2.2mm}
    \begin{tikzpicture}[spy using outlines={yellow,magnification=\ssmag,size=\ssizz},inner sep=0]
		\node {\includegraphics[width=\sswidth]{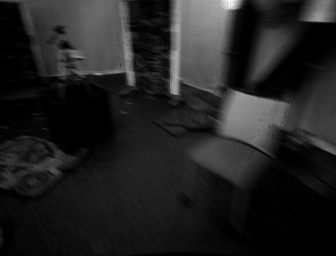}};
		\spy on \ssxxsthree in node [left] at \ssyystwo;
		\node [anchor=west] at \sccone {\textcolor{lime}{\bf E-CIR}};
	\end{tikzpicture} \hspace{-2.2mm}
	\begin{tikzpicture}[spy using outlines={yellow,magnification=\ssmag,size=\ssizz},inner sep=0]
		\node {\includegraphics[width=\sswidth]{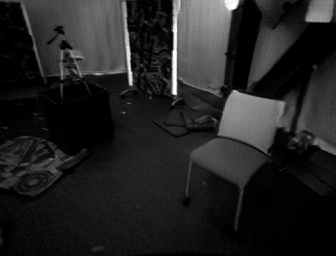}};
		\spy on \ssxxsthree in node [left] at \ssyystwo;
		\node [anchor=west] at \sccone {\textcolor{lime}{\bf \myname}};
	\end{tikzpicture} \vspace{0.5mm} \\
 
	\begin{tikzpicture}[spy using outlines={yellow,magnification=\ssmagtwo,size=\ssizz},inner sep=0]
		\node {\includegraphics[width=\sswidth]{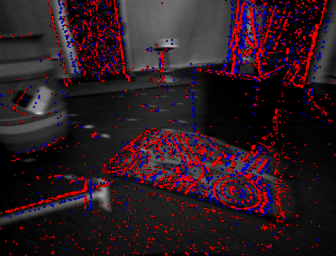}};
		\spy on \ssxxsone in node [right] at \ssyysone;
		\node [anchor=west] at \sccone {\textcolor{lime}{\bf Inputs}};
	\end{tikzpicture} \hspace{-2.2mm}
	\begin{tikzpicture}[spy using outlines={yellow,magnification=\ssmagtwo,size=\ssizz},inner sep=0]
		\node {\includegraphics[width=\sswidth]{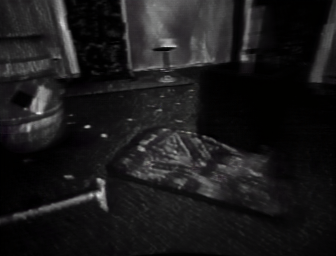}};
		\spy on \ssxxsone in node [right] at \ssyysone;
		\node [anchor=west] at \sccone {\textcolor{lime}{\bf DAVANet}};
	\end{tikzpicture} \hspace{-2.2mm}
	\begin{tikzpicture}[spy using outlines={yellow,magnification=\ssmagtwo,size=\ssizz},inner sep=0]
		\node {\includegraphics[width=\sswidth]{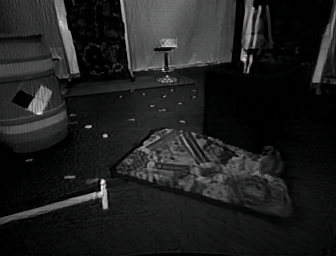}};
		\spy on \ssxxsone in node [right] at \ssyysone;
		\node [anchor=west] at \sccone {\textcolor{lime}{\bf LEVS}};
	\end{tikzpicture} \hspace{-2.2mm}
	\begin{tikzpicture}[spy using outlines={yellow,magnification=\ssmagtwo,size=\ssizz},inner sep=0]
		\node {\includegraphics[width=\sswidth]{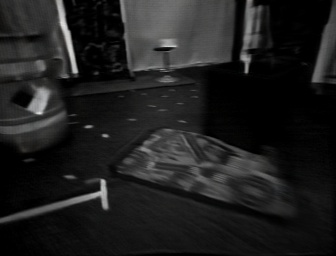}};
		\spy on \ssxxsone in node [right] at \ssyysone;
		\node [anchor=west] at \sccone {\textcolor{lime}{\bf Motion-ETR}};
	\end{tikzpicture} \hspace{-2.2mm}
	\begin{tikzpicture}[spy using outlines={yellow,magnification=\ssmagtwo,size=\ssizz},inner sep=0]
		\node {\includegraphics[width=\sswidth]{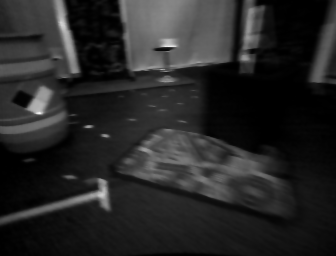}};
		\spy on \ssxxsone in node [right] at \ssyysone;
		\node [anchor=west] at \sccone {\textcolor{lime}{\bf EDI}};
	\end{tikzpicture} \vspace{0.25mm} \\

	\begin{tikzpicture}[spy using outlines={yellow,magnification=\ssmagtwo,size=\ssizz},inner sep=0]
		\node {\includegraphics[width=\sswidth]{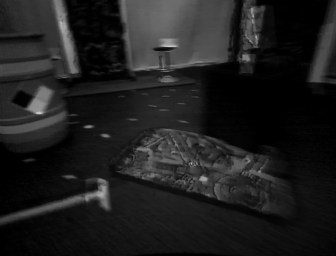}};
		\spy on \ssxxsone in node [right] at \ssyysone;
		\node [anchor=west] at \sccone {\textcolor{lime}{\bf eSL-Net}};
	\end{tikzpicture} \hspace{-2.2mm}
	\begin{tikzpicture}[spy using outlines={yellow,magnification=\ssmagtwo,size=\ssizz},inner sep=0]
		\node {\includegraphics[width=\sswidth]{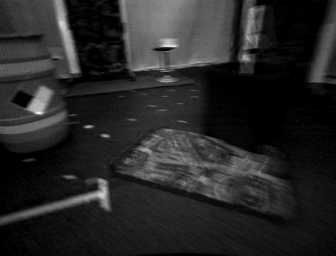}};
		\spy on \ssxxsone in node [right] at \ssyysone;
		\node [anchor=west] at \sccone {\textcolor{lime}{\bf LEDVDI}};
	\end{tikzpicture} \hspace{-2.2mm}
	\begin{tikzpicture}[spy using outlines={yellow,magnification=\ssmagtwo,size=\ssizz},inner sep=0]
		\node {\includegraphics[width=\sswidth]{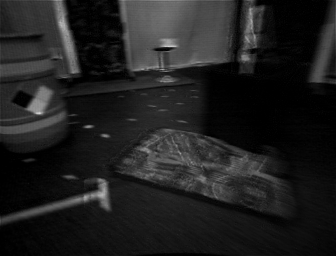}};
		\spy on \ssxxsone in node [right] at \ssyysone;
		\node [anchor=west] at \sccone {\textcolor{lime}{\bf RED-Net}};
	\end{tikzpicture} \hspace{-2.2mm}
	\begin{tikzpicture}[spy using outlines={yellow,magnification=\ssmagtwo,size=\ssizz},inner sep=0]
		\node {\includegraphics[width=\sswidth]{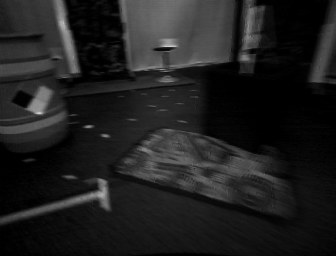}};
		\spy on \ssxxsone in node [right] at \ssyysone;
		\node [anchor=west] at \sccone {\textcolor{lime}{\bf E-CIR}};
	\end{tikzpicture} \hspace{-2.2mm}
	\begin{tikzpicture}[spy using outlines={yellow,magnification=\ssmagtwo,size=\ssizz},inner sep=0]
		\node {\includegraphics[width=\sswidth]{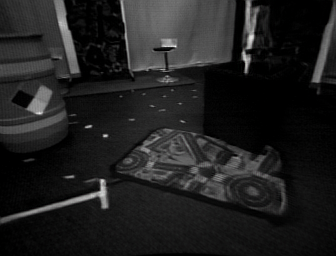}};
		\spy on \ssxxsone in node [right] at \ssyysone;
		\node [anchor=west] at \sccone {\textcolor{lime}{\bf \myname}};
	\end{tikzpicture} \vspace{0.5mm} \\
	\caption{Qualitative results of motion deblurring of 9 different methods on the \textit{MVSEC} dataset. We only select two exemplar frames for visualization.}
	\label{fig:qualitative_mvsec}
	\vspace{-4mm}
\end{figure*}

\begin{figure*}[t]
    \centering
    \includegraphics[width=\linewidth]{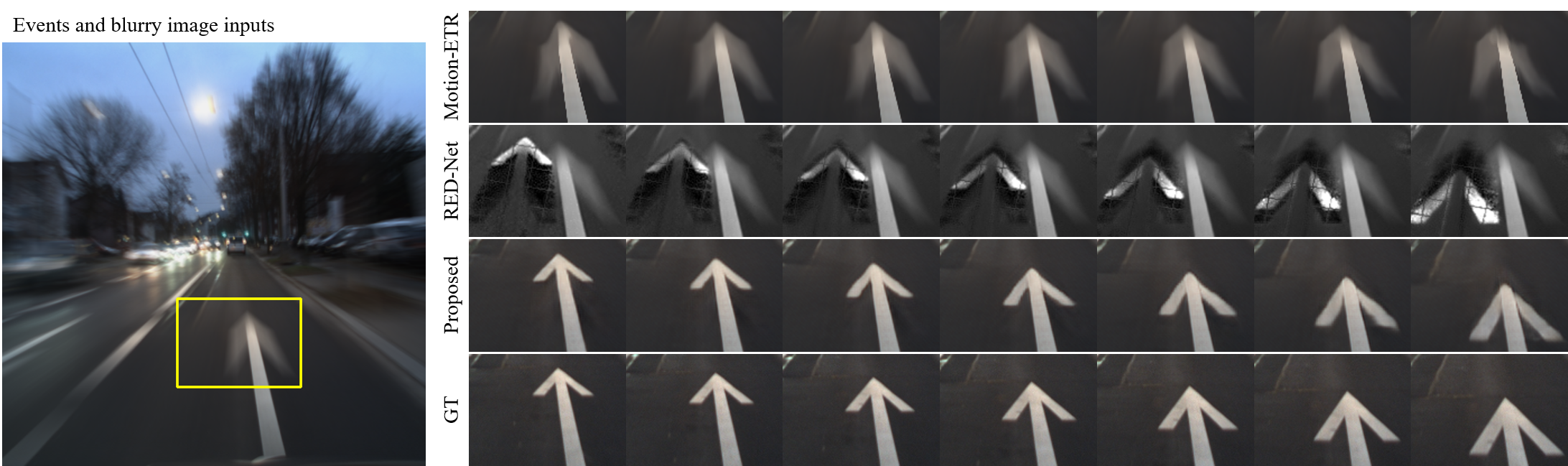}
    \caption{Multi-frame motion deblurring results on the \textit{DSEC-large} dataset.}
    \label{fig:multiframe}
    \vspace{-2.25mm}
\end{figure*}

\def\sswidth{0.247\textwidth}
\def\subwidth{0.1385\textwidth}
\def\sccone{(-2.2,1.45)}

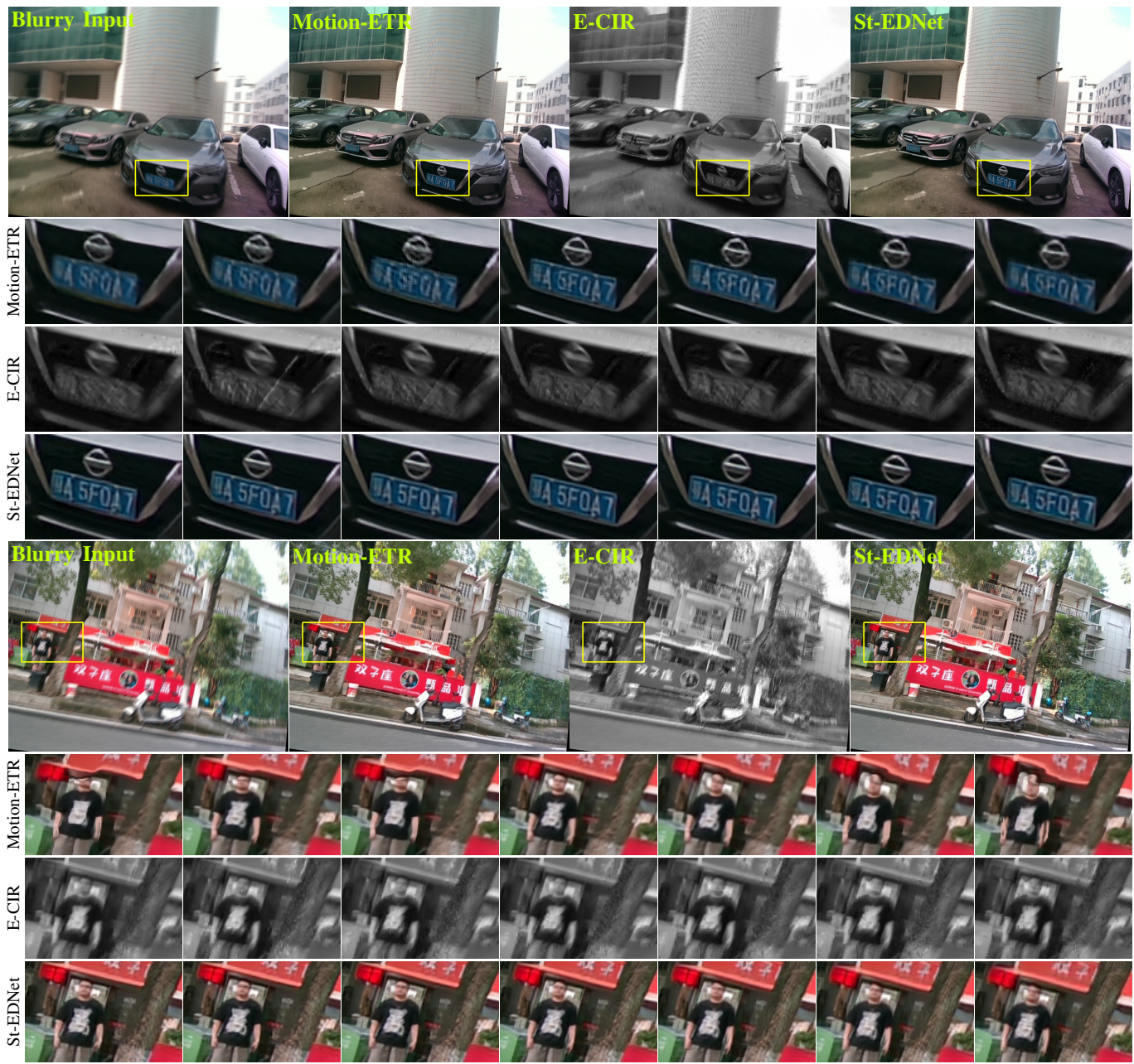
\begin{figure*}[t]
    \centering
    \begin{tikzpicture}[inner sep=0]
		\node {\includegraphics[width=\sswidth]{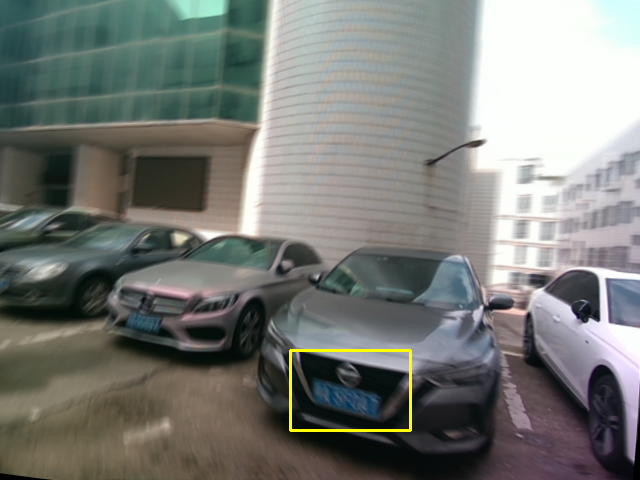}};
		\node [anchor=west] at \sccone {\textcolor{lime}{\bf Blurry Input}};
	\end{tikzpicture} \hspace{-2.2mm}
    \begin{tikzpicture}[inner sep=0]
		\node {\includegraphics[width=\sswidth]{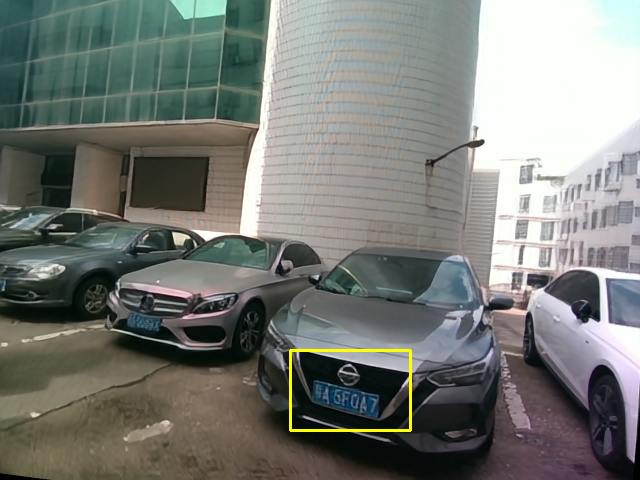}};
		\node [anchor=west] at \sccone {\textcolor{lime}{\bf Motion-ETR}};
	\end{tikzpicture} \hspace{-2.2mm}
    \begin{tikzpicture}[inner sep=0]
		\node {\includegraphics[width=\sswidth]{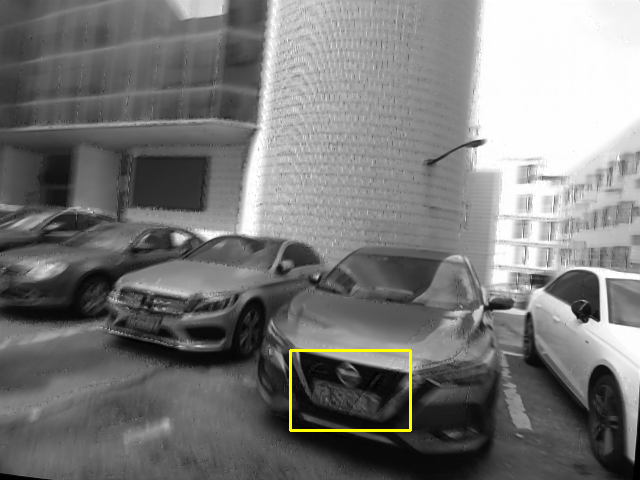}};
		\node [anchor=west] at \sccone {\textcolor{lime}{\bf E-CIR}};
	\end{tikzpicture} \hspace{-2.2mm}
    \begin{tikzpicture}[inner sep=0]
		\node {\includegraphics[width=\sswidth]{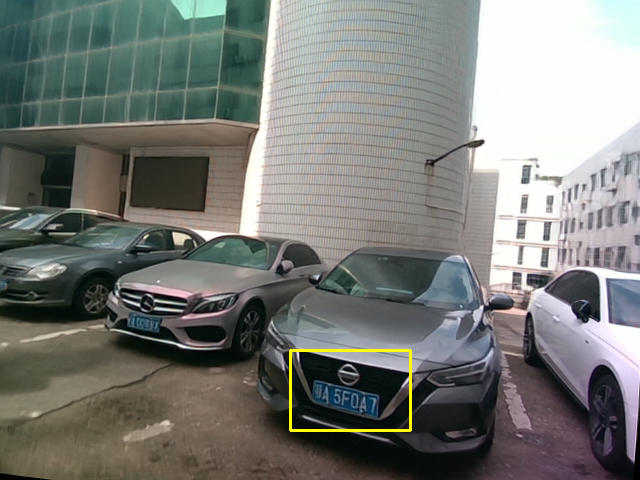}};
		\node [anchor=west] at \sccone {\textcolor{lime}{\bf St-EDNet}};
	\end{tikzpicture} \\
    \vspace{0.5mm}

    \begin{tikzpicture}[inner sep=0]
        \node {\includegraphics[width=\subwidth]{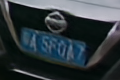}};
        \node [align=center, rotate=90] at (-1.45, 0) {\textcolor{black}{\footnotesize Motion-ETR}};
    \end{tikzpicture}  \hspace{-2.2mm}
    \begin{tikzpicture}[inner sep=0]
        \node {\includegraphics[width=\subwidth]{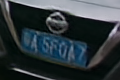}};
    \end{tikzpicture}  \hspace{-2.2mm}
    \begin{tikzpicture}[inner sep=0]
        \node {\includegraphics[width=\subwidth]{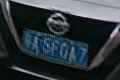}};
    \end{tikzpicture} \hspace{-2.2mm}
    \begin{tikzpicture}[inner sep=0]
        \node {\includegraphics[width=\subwidth]{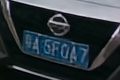}};
    \end{tikzpicture} \hspace{-2.2mm}
    \begin{tikzpicture}[inner sep=0]
        \node {\includegraphics[width=\subwidth]{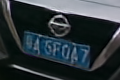}};
    \end{tikzpicture} \hspace{-2.2mm}
    \begin{tikzpicture}[inner sep=0]
        \node {\includegraphics[width=\subwidth]{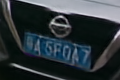}};
    \end{tikzpicture} \hspace{-2.2mm}
    \begin{tikzpicture}[inner sep=0]
        \node {\includegraphics[width=\subwidth]{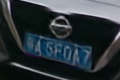}};
    \end{tikzpicture} \\
    \vspace{0.5mm}

    \begin{tikzpicture}[inner sep=0]
        \node {\includegraphics[width=\subwidth]{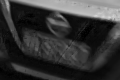}};
        \node [align=center, rotate=90] at (-1.45, 0) {\textcolor{black}{\footnotesize E-CIR}};
    \end{tikzpicture}  \hspace{-2.2mm}
    \begin{tikzpicture}[inner sep=0]
        \node {\includegraphics[width=\subwidth]{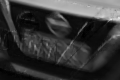}};
    \end{tikzpicture}  \hspace{-2.2mm}
    \begin{tikzpicture}[inner sep=0]
        \node {\includegraphics[width=\subwidth]{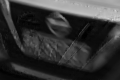}};
    \end{tikzpicture} \hspace{-2.2mm}
    \begin{tikzpicture}[inner sep=0]
        \node {\includegraphics[width=\subwidth]{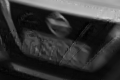}};
    \end{tikzpicture} \hspace{-2.2mm}
    \begin{tikzpicture}[inner sep=0]
        \node {\includegraphics[width=\subwidth]{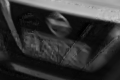}};
    \end{tikzpicture} \hspace{-2.2mm}
    \begin{tikzpicture}[inner sep=0]
        \node {\includegraphics[width=\subwidth]{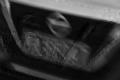}};
    \end{tikzpicture} \hspace{-2.2mm}
    \begin{tikzpicture}[inner sep=0]
        \node {\includegraphics[width=\subwidth]{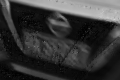}};
    \end{tikzpicture} \\
    \vspace{0.5mm}

    \begin{tikzpicture}[inner sep=0]
        \node {\includegraphics[width=\subwidth]{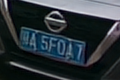}};
        \node [align=center, rotate=90] at (-1.45, 0) {\textcolor{black}{\footnotesize St-EDNet}};
    \end{tikzpicture}  \hspace{-2.2mm}
    \begin{tikzpicture}[inner sep=0]
        \node {\includegraphics[width=\subwidth]{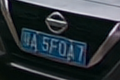}};
    \end{tikzpicture}  \hspace{-2.2mm}
    \begin{tikzpicture}[inner sep=0]
        \node {\includegraphics[width=\subwidth]{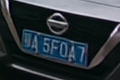}};
    \end{tikzpicture} \hspace{-2.2mm}
    \begin{tikzpicture}[inner sep=0]
        \node {\includegraphics[width=\subwidth]{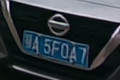}};
    \end{tikzpicture} \hspace{-2.2mm}
    \begin{tikzpicture}[inner sep=0]
        \node {\includegraphics[width=\subwidth]{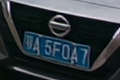}};
    \end{tikzpicture} \hspace{-2.2mm}
    \begin{tikzpicture}[inner sep=0]
        \node {\includegraphics[width=\subwidth]{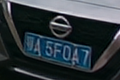}};
    \end{tikzpicture} \hspace{-2.2mm}
    \begin{tikzpicture}[inner sep=0]
        \node {\includegraphics[width=\subwidth]{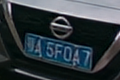}};
    \end{tikzpicture} \\
    \vspace{0.5mm}

    \begin{tikzpicture}[inner sep=0]
		\node {\includegraphics[width=\sswidth]{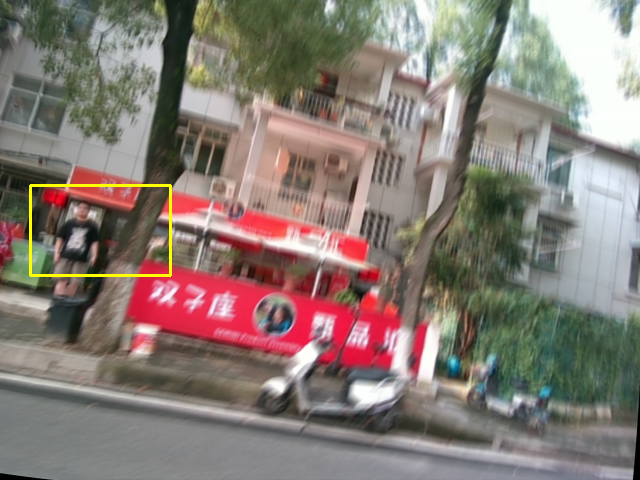}};
		\node [anchor=west] at \sccone {\textcolor{lime}{\bf Blurry Input}};
	\end{tikzpicture} \hspace{-2.2mm}
    \begin{tikzpicture}[inner sep=0]
		\node {\includegraphics[width=\sswidth]{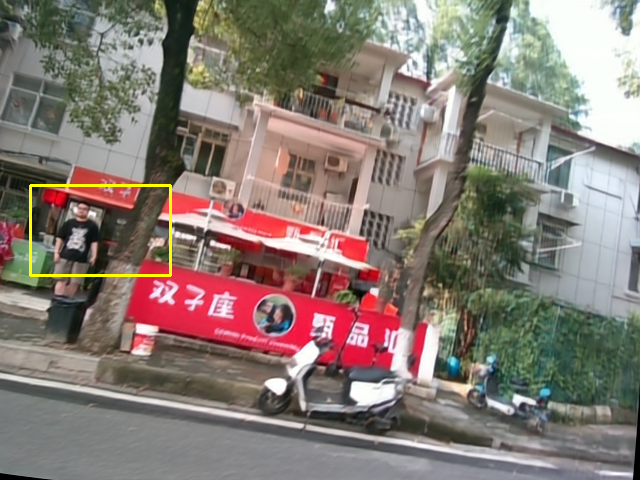}};
		\node [anchor=west] at \sccone {\textcolor{lime}{\bf Motion-ETR}};
	\end{tikzpicture} \hspace{-2.2mm}
    \begin{tikzpicture}[inner sep=0]
		\node {\includegraphics[width=\sswidth]{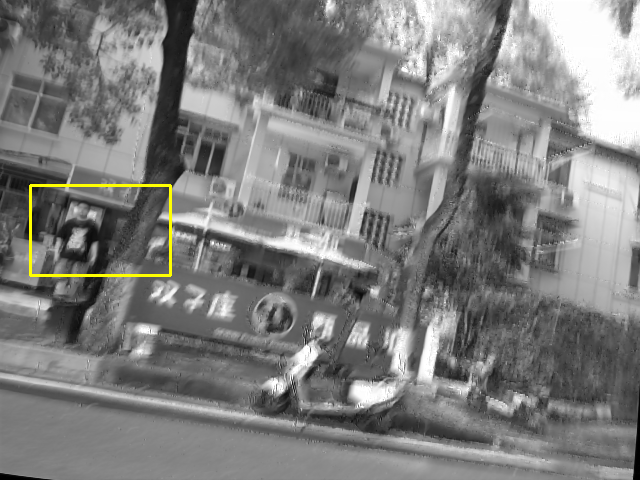}};
		\node [anchor=west] at \sccone {\textcolor{lime}{\bf E-CIR}};
	\end{tikzpicture} \hspace{-2.2mm}
    \begin{tikzpicture}[inner sep=0]
		\node {\includegraphics[width=\sswidth]{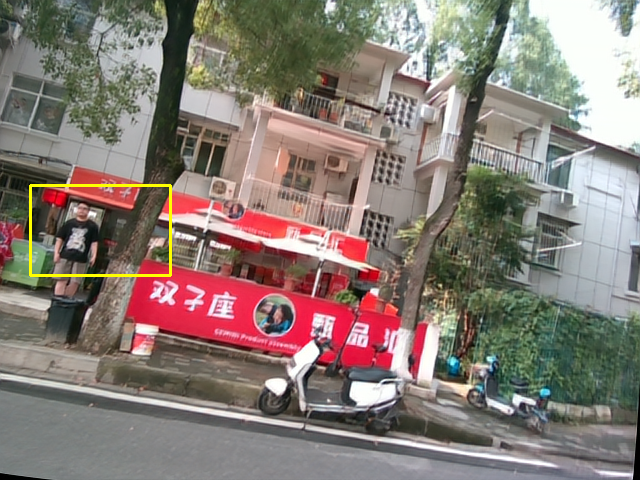}};
		\node [anchor=west] at \sccone {\textcolor{lime}{\bf St-EDNet}};
	\end{tikzpicture} \\
    \vspace{0.5mm}

    \begin{tikzpicture}[inner sep=0]
        \node {\includegraphics[width=\subwidth]{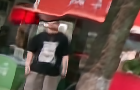}};
        \node [align=center, rotate=90] at (-1.45, 0) {\textcolor{black}{\footnotesize Motion-ETR}};
    \end{tikzpicture}  \hspace{-2.2mm}
    \begin{tikzpicture}[inner sep=0]
        \node {\includegraphics[width=\subwidth]{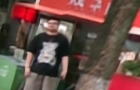}};
    \end{tikzpicture}  \hspace{-2.2mm}
    \begin{tikzpicture}[inner sep=0]
        \node {\includegraphics[width=\subwidth]{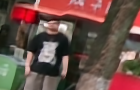}};
    \end{tikzpicture} \hspace{-2.2mm}
    \begin{tikzpicture}[inner sep=0]
        \node {\includegraphics[width=\subwidth]{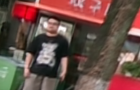}};
    \end{tikzpicture} \hspace{-2.2mm}
    \begin{tikzpicture}[inner sep=0]
        \node {\includegraphics[width=\subwidth]{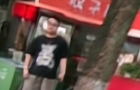}};
    \end{tikzpicture} \hspace{-2.2mm}
    \begin{tikzpicture}[inner sep=0]
        \node {\includegraphics[width=\subwidth]{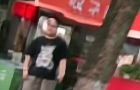}};
    \end{tikzpicture} \hspace{-2.2mm}
    \begin{tikzpicture}[inner sep=0]
        \node {\includegraphics[width=\subwidth]{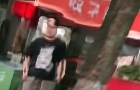}};
    \end{tikzpicture} \\
    \vspace{0.5mm}

    \begin{tikzpicture}[inner sep=0]
        \node {\includegraphics[width=\subwidth]{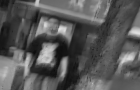}};
        \node [align=center, rotate=90] at (-1.45, 0) {\textcolor{black}{\footnotesize E-CIR}};
    \end{tikzpicture}  \hspace{-2.2mm}
    \begin{tikzpicture}[inner sep=0]
        \node {\includegraphics[width=\subwidth]{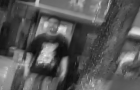}};
    \end{tikzpicture}  \hspace{-2.2mm}
    \begin{tikzpicture}[inner sep=0]
        \node {\includegraphics[width=\subwidth]{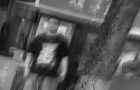}};
    \end{tikzpicture} \hspace{-2.2mm}
    \begin{tikzpicture}[inner sep=0]
        \node {\includegraphics[width=\subwidth]{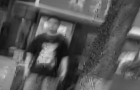}};
    \end{tikzpicture} \hspace{-2.2mm}
    \begin{tikzpicture}[inner sep=0]
        \node {\includegraphics[width=\subwidth]{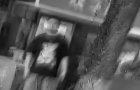}};
    \end{tikzpicture} \hspace{-2.2mm}
    \begin{tikzpicture}[inner sep=0]
        \node {\includegraphics[width=\subwidth]{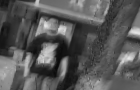}};
    \end{tikzpicture} \hspace{-2.2mm}
    \begin{tikzpicture}[inner sep=0]
        \node {\includegraphics[width=\subwidth]{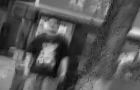}};
    \end{tikzpicture} \\
    \vspace{0.5mm}

    \begin{tikzpicture}[inner sep=0]
        \node {\includegraphics[width=\subwidth]{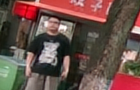}};
        \node [align=center, rotate=90] at (-1.45, 0) {\textcolor{black}{\footnotesize St-EDNet}};
    \end{tikzpicture}  \hspace{-2.2mm}
    \begin{tikzpicture}[inner sep=0]
        \node {\includegraphics[width=\subwidth]{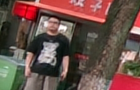}};
    \end{tikzpicture}  \hspace{-2.2mm}
    \begin{tikzpicture}[inner sep=0]
        \node {\includegraphics[width=\subwidth]{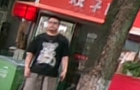}};
    \end{tikzpicture} \hspace{-2.2mm}
    \begin{tikzpicture}[inner sep=0]
        \node {\includegraphics[width=\subwidth]{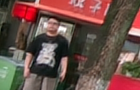}};
    \end{tikzpicture} \hspace{-2.2mm}
    \begin{tikzpicture}[inner sep=0]
        \node {\includegraphics[width=\subwidth]{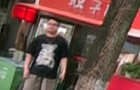}};
    \end{tikzpicture} \hspace{-2.2mm}
    \begin{tikzpicture}[inner sep=0]
        \node {\includegraphics[width=\subwidth]{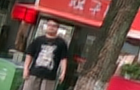}};
    \end{tikzpicture} \hspace{-2.2mm}
    \begin{tikzpicture}[inner sep=0]
        \node {\includegraphics[width=\subwidth]{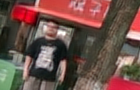}};
    \end{tikzpicture} \\

    \caption{Qualitative comparisons of the multi-frame motion deblurring on the \textit{\mydata} dataset.}
    \label{fig:qualitative_steic}
\end{figure*}

\begin{table*}[t]
    \centering
    \caption{Quantitative comparisons on the motion deblurring task of the proposed \myname\ to the state-of-the-art methods on the \textit{DSEC-large}, the \textit{MVSEC} and our \textit{\mydata} datasets. Note that for the single frame prediction, all methods are evaluated with the middle frame of the sequence predictions except for eSL-Net, which uses the first frame according to \cite{wang2020event}. For the sequence prediction, LEDVDI only outputs 6 frames, while the others output 7 frames.}
    \vspace{-2mm}
    \begin{tabular}{l c p{1.5cm}<{\centering} p{1.5cm}<{\centering} p{1.5cm}<{\centering} c p{1.5cm}<{\centering} p{1.5cm}<{\centering} p{1.5cm}<{\centering}}
        \toprule[1.pt]
        \multirow{2}{*}{Method} & \multirow{2}{*}{isColor} & \multicolumn{3}{c}{Single frame prediction} & & \multicolumn{3}{c}{Sequence prediction} \\
        \cline{3-5} \cline{7-9}
        & & PSNR$\uparrow$ & SSIM$\uparrow$ & LPIPS$\downarrow$ & & PSNR$\uparrow$ & SSIM$\uparrow$ & LPIPS$\downarrow$ \\
        \hline
        \textbf{\textit{DSEC-large}}\\
        \hline
        DAVANet \cite{zhou2019davanet} & \ding{52} & 24.557 & 0.7734 & 0.2601 && / & / & / \\
        LEVS \cite{jin2018learning} & \ding{52} & 25.340 & 0.7721 & 0.1856 && 23.097 & 0.7248 & 0.2205 \\
        Motion-ETR \cite{zhang2021exposure} & \ding{52} & 26.176 & 0.7983 & 0.2283 && 23.120 & 0.7147 & 0.2498 \\
        EDI \cite{pan2019bringing} & \ding{56} & 23.733 & 0.7498 & 0.2969 && 22.545 & 0.7157 & 0.3004 \\
        eSL-Net \cite{wang2020event} & \ding{56} & 19.004 & 0.5741 & 0.2843 && 17.671 & 0.5451 & 0.3185 \\
        LEDVDI \cite{lin2020learning} & \ding{56} & 21.479 & 0.6828 & 0.2256 && 20.368 & 0.6570 & 0.2424 \\
        RED-Net \cite{xu2021motion} & \ding{56} & 21.367 & 0.6354 & 0.2385 && 19.841 & 0.5832 & 0.2850 \\
        RED-Net$^*$ & \ding{56} & 27.242 & 0.8287 & 0.1577 && 27.219 & 0.8313 & 0.1486 \\
        E-CIR \cite{song2022cir} & \ding{56} & 20.729 & 0.6323 & 0.3007 && 19.837 & 0.5920 & 0.3205 \\
        E-CIR$^*$ & \ding{56} & 21.460 & 0.6785 & 0.2693 && 21.089 & 0.6603 & 0.2772 \\
        \myname\ (Ours) & \ding{52} & \textbf{29.065} & \textbf{0.8738} & \textbf{0.1033} && \textbf{28.995} & \textbf{0.8759} & \textbf{0.0994} \\
        \hline
        \textbf{\textit{MVSEC}}\\
        \hline
        DAVANet \cite{zhou2019davanet} & \ding{52} & 23.957 & 0.6918 & 0.2578 && / & / & / \\
        LEVS \cite{jin2018learning} & \ding{52} & 28.346 & 0.8038 & 0.1392 && 24.882 & 0.6991 & 0.1829 \\
        Motion-ETR \cite{zhang2021exposure} & \ding{52} & 27.229 & 0.7789 & 0.2125 && 23.079 & 0.6163 & 0.2449 \\
        EDI \cite{pan2019bringing} & \ding{56} & 25.427 & 0.7457 & 0.3044 && 24.086 & 0.6991 & 0.3110 \\
        eSL-Net \cite{wang2020event} & \ding{56} & 24.099 & 0.6696 & 0.2290 && 25.306 & 0.7190 & 0.2112 \\
        LEDVDI \cite{lin2020learning} & \ding{56} & 25.802 & 0.7751 & 0.1904 && 24.529 & 0.7376 & 0.2031 \\
        RED-Net \cite{xu2021motion} & \ding{56} & 27.122 & 0.7743 & 0.1877 && 25.461 & 0.7246 & 0.1975\\
        RED-Net$^*$ & \ding{56} & 29.875 & 0.8425 & 0.1078 && 29.618 & 0.8385 & 0.1078 \\
        E-CIR \cite{song2022cir} & \ding{56} & 27.495 & 0.7797 & 0.2214 && 25.804 & 0.7292 & 0.2307 \\
        E-CIR$^*$ & \ding{56} & 27.792 & 0.7872 & 0.2015 && 26.773 & 0.7575 & 0.2082 \\
        \myname\ (Ours) & \ding{52} & \textbf{30.129} & \textbf{0.8462} & \textbf{0.1016} && \textbf{29.951} & \textbf{0.8447} & \textbf{0.1020} \\
        \hline
        \textbf{\textit{StEIC}}\\
        \hline
        LEVS \cite{jin2018learning} & \ding{52} & 24.660 & 0.8022 & 0.1598 && 21.186 & 0.6513 & 0.2155 \\
        Motion-ETR \cite{zhang2021exposure} & \ding{52} & 25.057 & 0.8198 & 0.2113 && 19.944 & 0.5945 & 0.2554 \\
        EDI \cite{pan2019bringing} & \ding{56} & 18.923 & 0.5972 & 0.3634 && 18.130 & 0.5445 & 0.3692 \\
        eSL-Net \cite{wang2020event} & \ding{56} & 16.662 & 0.4886 & 0.2995 && 16.563 & 0.4829 & 0.3093 \\
        LEDVDI \cite{lin2020learning} & \ding{56} & 20.764 & 0.6391 & 0.2244 && 19.410 & 0.5895 & 0.2427 \\
        RED-Net \cite{xu2021motion} & \ding{56} & 20.198 & 0.6241 & 0.2359 && 18.506 & 0.5615 & 0.2642 \\
        RED-Net$^*$ & \ding{56} & 26.511 & 0.8610 & 0.0994 && 26.202 & 0.8545 & 0.0977 \\
        E-CIR \cite{song2022cir} & \ding{56} & 22.004 & 0.6688 & 0.2883 && 20.441 & 0.6059 & 0.2933 \\
        E-CIR$^*$ & \ding{56} & 22.076 & 0.6720 & 0.2837 && 20.706 & 0.6166 & 0.2896 \\
        \myname\ (Ours) & \ding{52} & \textbf{26.724} & \textbf{0.8707} & \textbf{0.0981} && \textbf{26.344} & \textbf{0.8633} & \textbf{0.0976} \\
        \toprule[1.pt]
    \end{tabular}
    \label{tab:comparison}
\end{table*}

\section{StEIC Dataset}
Existing datasets \cite{zhu2018multivehicle, gehrig2021dsec} use LiDAR to capture ground-truth depth with low frame rate and sparsity. To obtain denser depth ground truth, they introduce offline mapping methods, \eg, LIO-Mapping \cite{ye2019tightly}, which rely on static scene assumptions. This causes the depth of moving objects, like the car in \cref{fig:coarsedisp}, to be filtered out as outliers, leaving holes without valid ground-truth depth. It motivates us to build a dataset containing intensity images, events, and dense depth maps with STereo Event and Intensity Cameras (\textit{\mydata}). 

We build a stereo hybrid camera system composed of a SilkyEvCam event camera with the Prophesee Gen3.1 (VGA) sensor and an Intel RealSense D455 RGB-D camera (\cref{fig:camera_cam}), both two cameras sharing the same resolution of 640$\times$480. With the D455 RGB-D camera, our \textit{\mydata} provides the temporally synchronized intensity images and dense ground-truth depths at a higher frame rate (60 FPS) than the abovementioned datasets as shown in \cref{tab:dataset}. We capture data with the Robot Operating System (ROS), using the host computer's time to synchronize the event and RGB-D cameras and manually fine-tune the timestamps. For spatial calibration, we first utilize the event-based image reconstruction method E2VID \cite{rebecq2019events} to translate event streams into intensity images and then use a checkerboard pattern together with the MATLAB camera calibration toolbox to estimate the intrinsic and extrinsic parameters of stereo cameras. 

As shown in \cref{fig:camera_scene}, our \textit{\mydata} records real-world street scenarios, including on-street parking, sidewalks, community areas, and diverse objects such as driving vehicles, pedestrians, electric bicycles, and buildings. As summarized in \cref{tab:dataset2}, \textit{\mydata} consists of 65 sequences with 5661 blurry-sharp pairs and corresponding event streams, including on-street parking, sidewalks, and community areas.

\section{Experiments}
\subsection{Datasets and Training Details}
\subsubsection{Datasets}
Our proposed \myname\ is trained on three real-world datasets: the Multi Vehicle Stereo Event Camera (\textit{MVSEC}) dataset \cite{zhu2018multivehicle}, the Stereo Event Camera dataset for Driving Scenarios (\textit{DSEC}) \cite{gehrig2021dsec}, and our proposed STereo Event and Intensity Cameras dataset (\textit{\mydata}).



\noindent\textbf{\textit{MVSEC}.} We select 3 sequences on the “indoor\_flying\_data” scene as the training set and the other 1 sequence as the testing set. Training and testing sets have 960 and 442 blurry-sharp pairs, respectively. We do camera and stereo rectification with the calibration parameters provided for the events and images.

\noindent\textbf{\textit{DSEC}.} Excluding 8 low-illumination night sequences, we use the other 23 sequences as the training set and 10 as the testing set. There are 4982 and 1851 blurry-sharp pairs in the training and testing set, respectively. We pair the images captured from the left intensity camera “cam1” with events captured from the right event camera “cam3” with the baseline of 0.6 meters to form the \textit{DSEC-large} dataset. As the calibration parameters and rectified images are provided, we apply camera rectification for the event cameras and stereo rectification as the regular preprocessing for the stereo-matching task. 

\noindent\textbf{\textit{\mydata}.} We divide \textit{\mydata} into two parts, \ie, 43 sequences as the training set and 22 sequences as the testing set. The training and testing set has 3702 and 1959 blurry-sharp pairs, respectively. Using the calibration parameters, we rectify events, intensity images, and disparities using the same methods described above.

The blurry images are synthesized to simulate long-exposure and high-speed motion situations in our experiments. We first increase the frame rate by interpolating 7 images between consecutive frames with RIFE \cite{huang2022real} and then generate blurry images based on the interpolated high frame-rate sequences. Then, blurry images are obtained by averaging over 49 consecutive images.

\subsubsection{Training Details} 
Our network is implemented using Pytorch and trained on a single NVIDIA Geforce RTX 3090 GPU with batch size 6 by default. During training, we randomly crop the samples into $256 \times 256$ patches. Adam optimizer is used, and the maximum epoch of training iterations is set to 120. The learning rate starts at $10^{-4}$, then decays by 50\% every 20 epochs from the 40th epoch. In our implementation, the number $N$ of DDFE modules is 6. The number $M$ of deblurred images $\{\mathbf{I}_m\}^{M-1}_{m=0}$ is 7. The hyperparameters $\{\lambda_1, \lambda_2, \lambda_3\}$ are set as: $\{1, 0.002, 0.0005\}$. All input blurry images and the concurrent events are temporally calibrated before feeding into the network.

\subsection{Results of Motion Deblurring}
We compare the proposed \myname\ with state-of-the-art methods including two conventional video deblurring methods, \ie, LEVS \cite{jin2018learning} and Motion-ETR \cite{zhang2021exposure}, one conventional stereo deblurring method, \ie, DAVANet \cite{zhou2019davanet}, and five recent event-based motion deblurring methods, \ie, EDI \cite{pan2019bringing}, eSL-Net \cite{wang2020event}, LEDVDI \cite{lin2020learning}, RED-Net \cite{xu2021motion}, and E-CIR \cite{song2022cir}. 

\subsubsection{Qualitative Comparisons} The qualitative results of the single frame reconstruction task are shown in \cref{fig:qualitative_dsec,fig:qualitative_mvsec}, where we select two exemplar restorations, respectively, from the \textit{DSEC-large} and \textit{MVSEC} datasets. The proposed \myname\ achieves the best visual performance among all state-of-the-art methods and gives sharp edges, smooth surfaces, and clear textures. Without the help of events, LEVS and Motion-ETR fail to recover the latent information, \eg, the traffic light. For event-based deblurring methods, the competitors cannot effectively deal with severe parallax between the intensity image and events and thus suffer from serious artifacts and missing details. Furthermore, in \cref{fig:multiframe,fig:qualitative_steic}, we compare the multi-frame output of our method with a frame-based method Motion-ETR and two event-based methods RED-Net and E-CIR on the \textit{DSEC-large} and the \textit{\mydata} dataset. \myname\ can reconstruct sequences of sharp and clear images, while Motion-ETR is limited in producing deblurred results, and event-based deblurring methods generate serious artifacts.

\subsubsection{Quantitative Comparisons} The quantitative results on a single frame and sequence image reconstructions are given in \cref{tab:comparison}. Overall, our proposed \myname\ outperforms the state-of-the-art methods by a large margin on the \textit{DSEC}, \textit{MVSEC}, and \textit{\mydata} datasets in terms of Peak Signal to Noise Ratio (PSNR, higher is better), Structural SIMilarity (SSIM, higher is better) \cite{wang2004image}, and Learned Perceptual Image Patch Similarity (LPIPS, lower is better) \cite{zhang2018unreasonable}. 

We first compare the conventional intensity-based methods, \ie, LEVS and Motion-ETR, and state-of-the-art event-based methods, \ie, EDI, eSL-Net, LEDVDI, RED-Net, and E-CIR. As shown in \cref{fig:qualitative_dsec,fig:qualitative_mvsec} and Tab.~\ref{tab:comparison}, LEVS and motion-ETR do not have the artifacts generated by the misaligned inputs and achieve relatively higher PSNR and SSIM, and lower LPIPS than state-of-the-art event-based methods for single frame prediction. Specifically, EDI fails to reconstruct sharp contours and clear textures as shown \cref{fig:qualitative_dsec,fig:qualitative_mvsec}, due to the invalid motion embedded in misaligned events. Even though the learning-based techniques, \ie, eSL-Net, LEDVDI, RED-Net, and E-CIR, can give sharper results than EDI, serious artifacts are produced as shown in \cref{fig:qualitative_dsec,fig:qualitative_mvsec}. Regarding sequence prediction, the performance gap between intensity-based and event-based methods is reduced remarkably due to the lack of intra-frame motion information for intensity-based methods. By considering the parallax between events and frames, our \myname\ outperforms both intensity-based and event-based state-of-the-art approaches, which validates the importance of alignment of events and frames for event-based methods.

We then compare \myname\ with the intensity-based stereo deblurring method, \ie, DAVANet. To have a similar setting as its original paper, we feed DAVANet with the stereo intensity data from the left intensity camera “cam1” and the right one “cam2”, whose images are warped to align the field of view of the right event camera “cam3”, on the \textit{DSEC-large} dataset. On the \textit{MVSEC} dataset, we implement DAVANet with the input stereo intensity data from the “left” and “right” cameras. Qualitative and quantitative comparisons show that \myname\ outperforms the intensity-based stereo deblurring method by a large margin in terms of PSNR, SSIM, and LPIPS, demonstrating that the aligned events can provide missing intra-frame information and boost deblurring performance.

For the sake of fair comparisons, we employ the recently proposed networks, \ie, RED-Net and E-CIR, and retrain them on the same datasets with full supervision as ours, and we denote them as RED-Net$^*$ and E-CIR$^*$. Compared to the officially provided weights, RED-Net$^*$ and E-CIR$^*$ perform better and show adaptability to the datasets. However, due to the parallax between the events and the frames, our \myname\ still outperforms the retrained RED-Net$^*$ and E-CIR$^*$, which validates the superiority of our proposed network.

\def\ssxxsone{(0.,0.44)} 
\def\ssyysone{(1.37,-0.58)} 
\def\ssxxstwo{(-1.2,.86)} 
\def\ssyystwo{(1.37,-0.58)} 
\def\ssizz{0.9cm} 
\def\sswidth{0.197\textwidth} 
\def\ssmag{3.5}
\def\scc{(-1.4,0.92)}
\def\sccone{(-1.75,1.12)}

\begin{figure*}[t]
	\centering
	\begin{tikzpicture}[spy using outlines={green,magnification=\ssmag,size=\ssizz},inner sep=0]
		\node {\includegraphics[width=\sswidth]{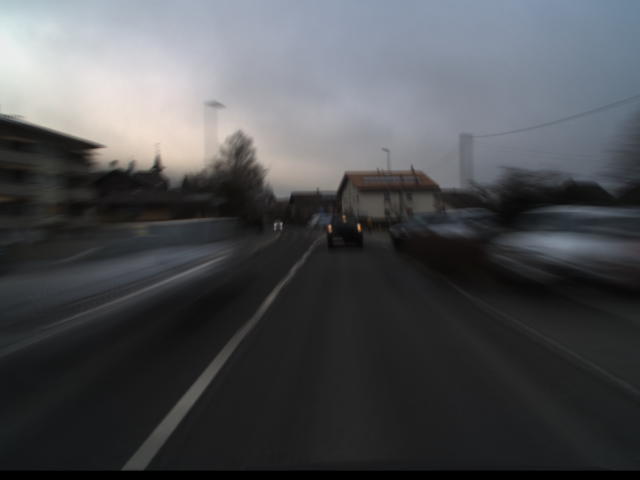}};
        \node [anchor=west] at \sccone {\textcolor{lime}{\bf Blurry image}};
	\end{tikzpicture} \hspace{-2.2mm}
		\begin{tikzpicture}[spy using outlines={green,magnification=\ssmag,size=\ssizz},inner sep=0]
		\node {\includegraphics[width=\sswidth]{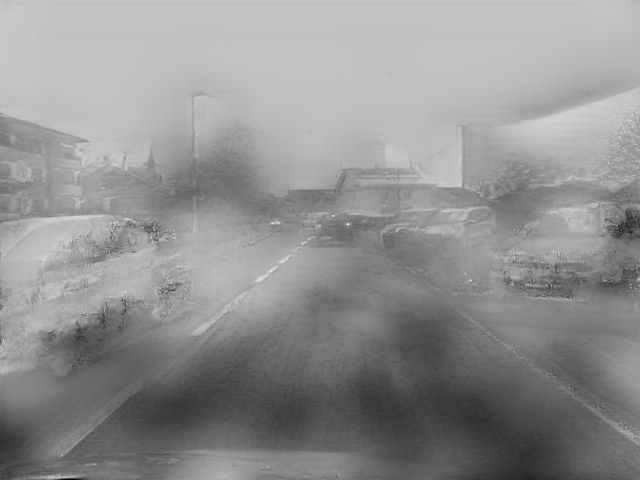}};
        \node [anchor=west] at \sccone {\textcolor{lime}{\bf E2VID}};
	\end{tikzpicture} \hspace{-2.2mm}
	\begin{tikzpicture}[spy using outlines={green,magnification=\ssmag,size=\ssizz},inner sep=0]
		\node {\includegraphics[width=\sswidth]{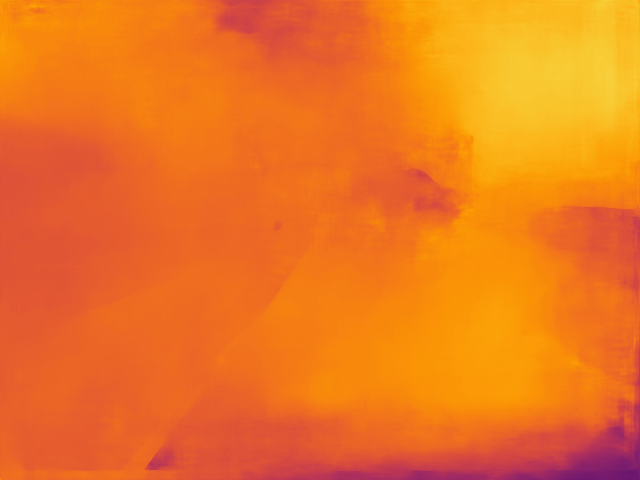}};
        \node [anchor=west] at \sccone {\textcolor{lime}{\bf E2VID+AANet}};
	\end{tikzpicture} \hspace{-2.2mm}
	\begin{tikzpicture}[spy using outlines={green,magnification=\ssmag,size=\ssizz},inner sep=0]
		\node {\includegraphics[width=\sswidth]{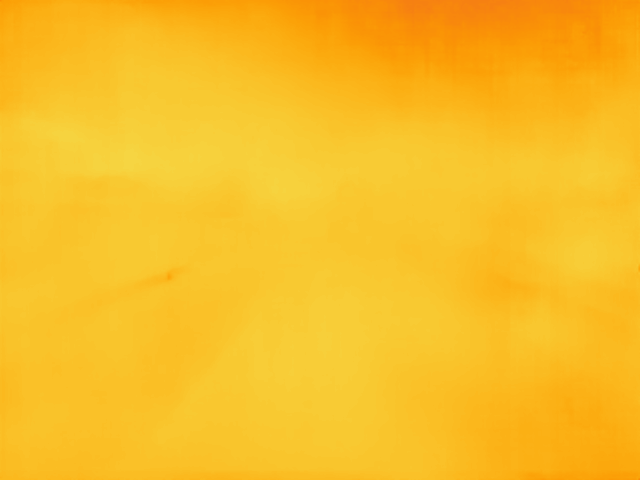}};
        \node [anchor=west] at \sccone {\textcolor{lime}{\bf E2VID+CFNet}};
	\end{tikzpicture} \hspace{-2.2mm}
		\begin{tikzpicture}[spy using outlines={green,magnification=\ssmag,size=\ssizz},inner sep=0]
		\node {\includegraphics[width=\sswidth]{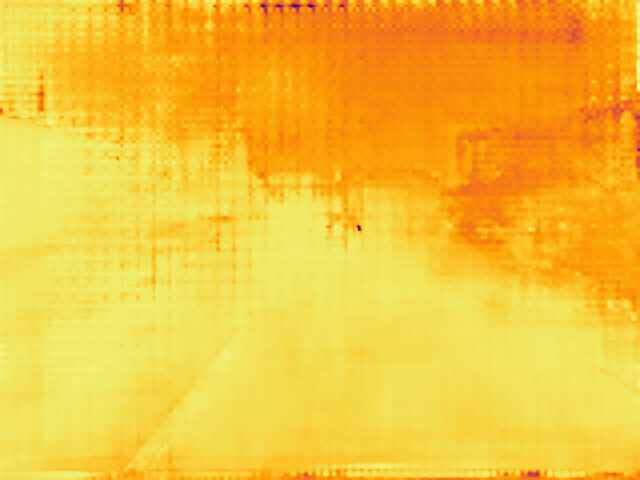}};
        \node [anchor=west] at \sccone {\textcolor{lime}{\bf E2VID+ACVNet}};
	\end{tikzpicture} \vspace{0.25mm} \\
	\begin{tikzpicture}[spy using outlines={green,magnification=\ssmag,size=\ssizz},inner sep=0]
		\node {\includegraphics[width=\sswidth]{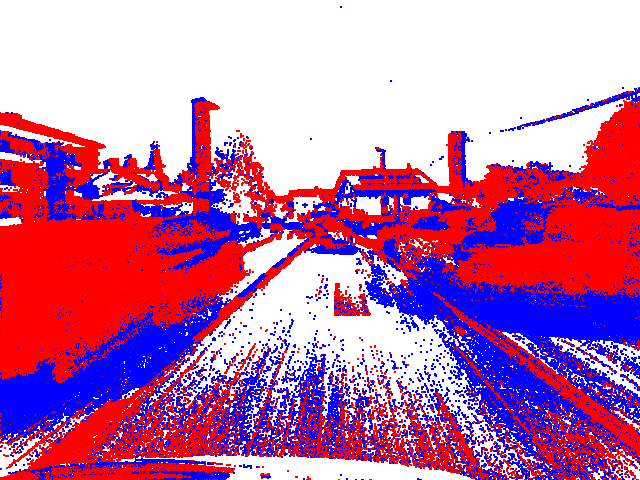}};
        \node [anchor=west] at \sccone {\textcolor{lime}{\bf Events}};
	\end{tikzpicture} \hspace{-2.2mm}
	\begin{tikzpicture}[spy using outlines={green,magnification=\ssmag,size=\ssizz},inner sep=0]
		\node {\includegraphics[width=\sswidth]{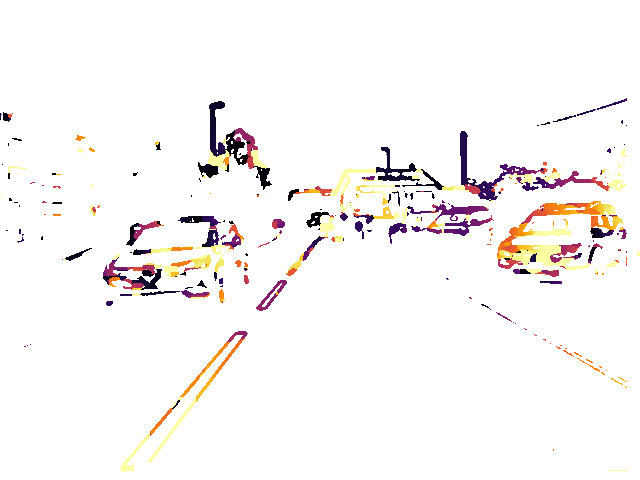}};
        \node [anchor=west] at \sccone {\textcolor{lime}{\bf HSM}};
	\end{tikzpicture} \hspace{-2.2mm}
	\begin{tikzpicture}[spy using outlines={green,magnification=\ssmag,size=\ssizz},inner sep=0]
		\node {\includegraphics[width=\sswidth]{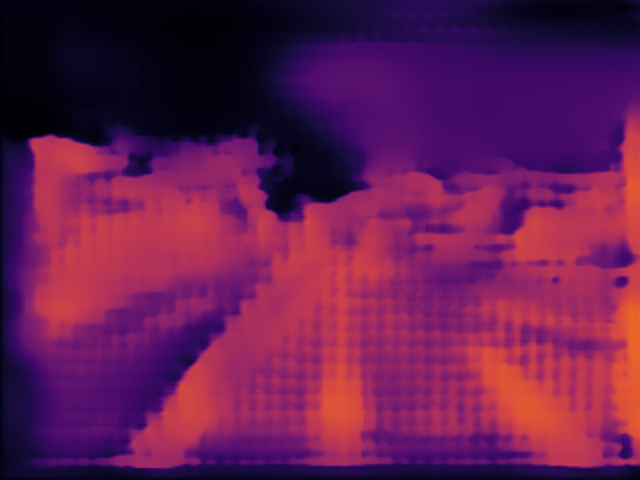}};
        \node [anchor=west] at \sccone {\textcolor{lime}{\bf SIES}};
	\end{tikzpicture} \hspace{-2.2mm}
		\begin{tikzpicture}[spy using outlines={green,magnification=\ssmag,size=\ssizz},inner sep=0]
		\node {\includegraphics[width=\sswidth]{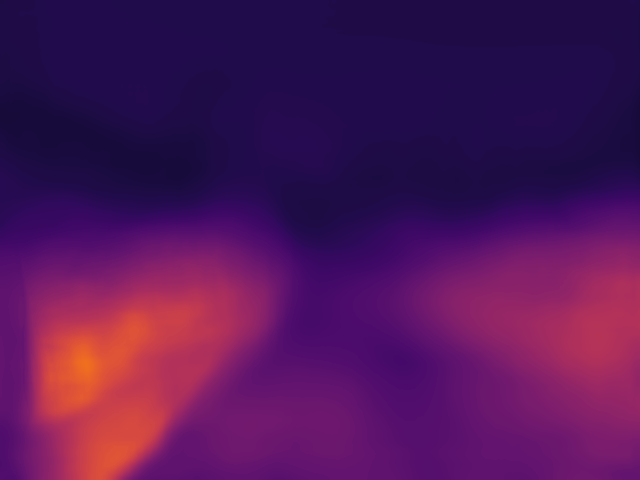}};
        \node [anchor=west] at \sccone {\textcolor{lime}{\bf Our $\mathbf{\hat{D}}_T$}};
	\end{tikzpicture} \hspace{-2.2mm}
		\begin{tikzpicture}[spy using outlines={green,magnification=\ssmag,size=\ssizz},inner sep=0]
		\node {\includegraphics[width=\sswidth]{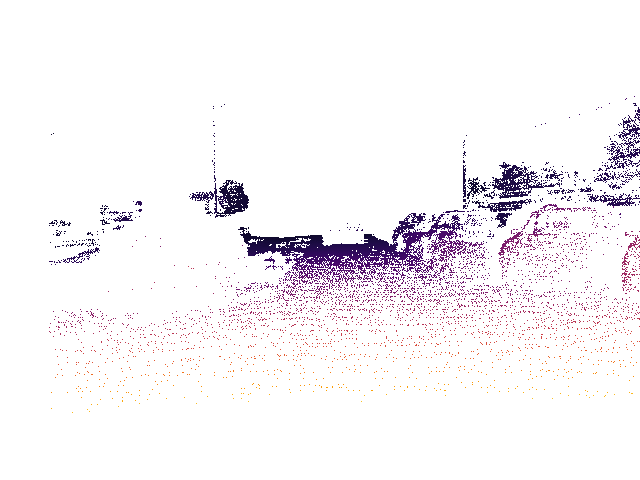}};
        \node [anchor=west] at \sccone {\textcolor{lime}{\bf GT}};
	\end{tikzpicture} \vspace{0.5mm} \\
    \begin{tikzpicture}[spy using outlines={green,magnification=\ssmag,size=\ssizz},inner sep=0]
		\node {\includegraphics[width=\sswidth]{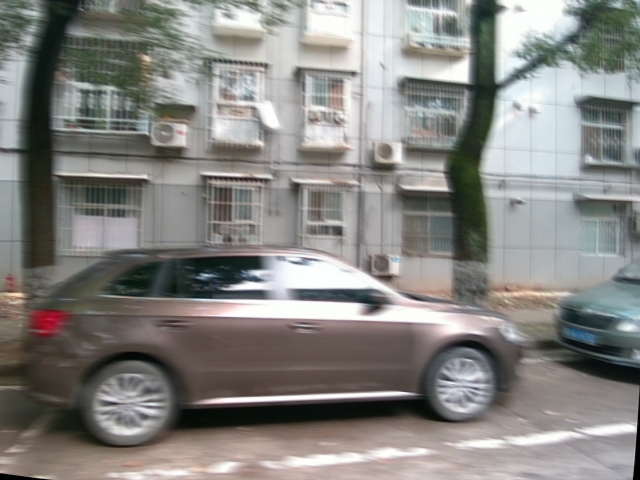}};
        \node [anchor=west] at \sccone {\textcolor{lime}{\bf Blurry image}};
	\end{tikzpicture} \hspace{-2.2mm}
		\begin{tikzpicture}[spy using outlines={green,magnification=\ssmag,size=\ssizz},inner sep=0]
		\node {\includegraphics[width=\sswidth]{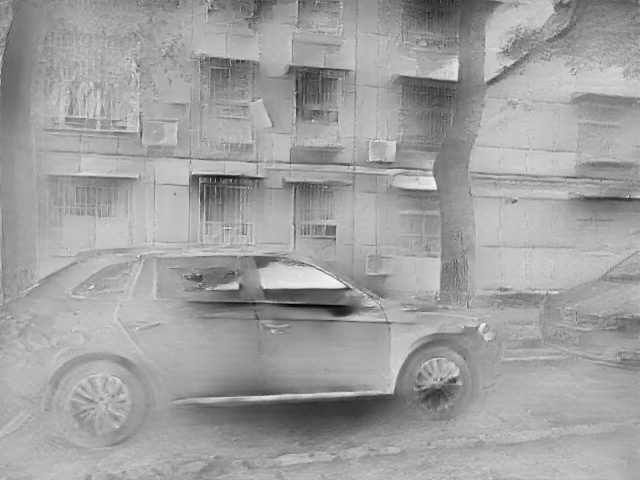}};
        \node [anchor=west] at \sccone {\textcolor{lime}{\bf E2VID}};
	\end{tikzpicture} \hspace{-2.2mm}
	\begin{tikzpicture}[spy using outlines={green,magnification=\ssmag,size=\ssizz},inner sep=0]
		\node {\includegraphics[width=\sswidth]{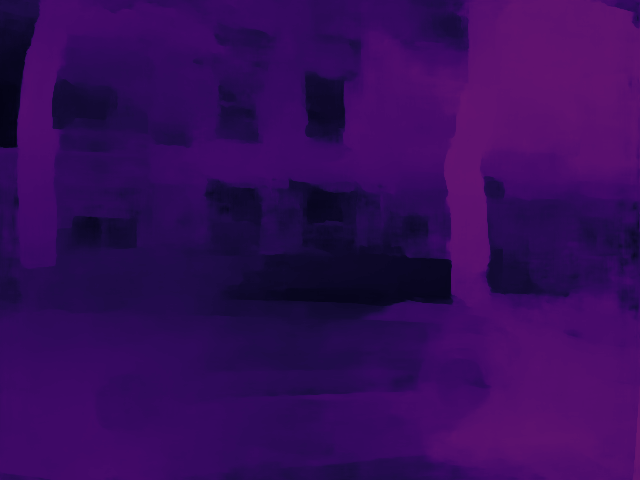}};
        \node [anchor=west] at \sccone {\textcolor{lime}{\bf E2VID+AANet}};
	\end{tikzpicture} \hspace{-2.2mm}
	\begin{tikzpicture}[spy using outlines={green,magnification=\ssmag,size=\ssizz},inner sep=0]
		\node {\includegraphics[width=\sswidth]{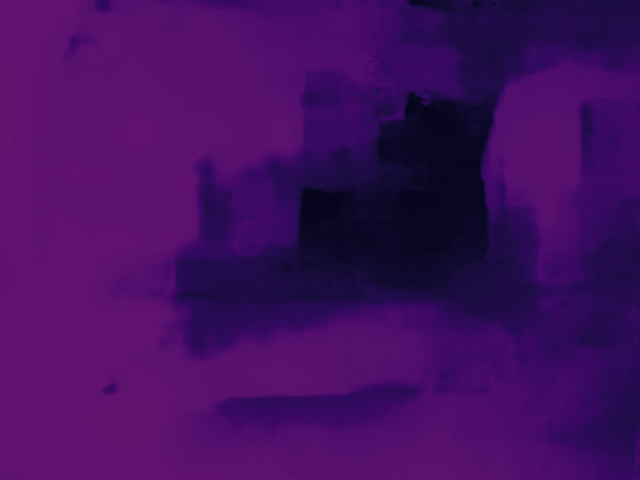}};
        \node [anchor=west] at \sccone {\textcolor{lime}{\bf E2VID+CFNet}};
	\end{tikzpicture} \hspace{-2.2mm}
		\begin{tikzpicture}[spy using outlines={green,magnification=\ssmag,size=\ssizz},inner sep=0]
		\node {\includegraphics[width=\sswidth]{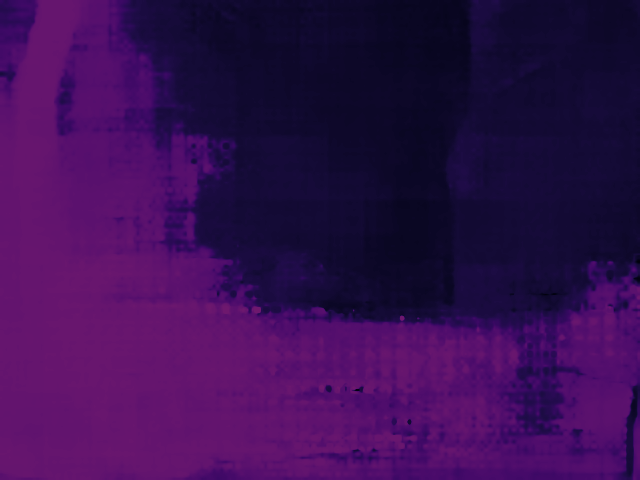}};
        \node [anchor=west] at \sccone {\textcolor{lime}{\bf E2VID+ACVNet}};
	\end{tikzpicture} \vspace{0.25mm} \\
	\begin{tikzpicture}[spy using outlines={green,magnification=\ssmag,size=\ssizz},inner sep=0]
		\node {\includegraphics[width=\sswidth]{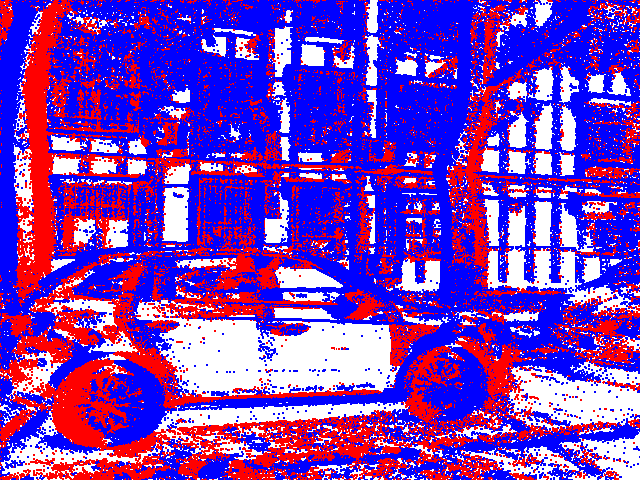}};
        \node [anchor=west] at \sccone {\textcolor{lime}{\bf Events}};
	\end{tikzpicture} \hspace{-2.2mm}
	\begin{tikzpicture}[spy using outlines={green,magnification=\ssmag,size=\ssizz},inner sep=0]
		\node {\includegraphics[width=\sswidth]{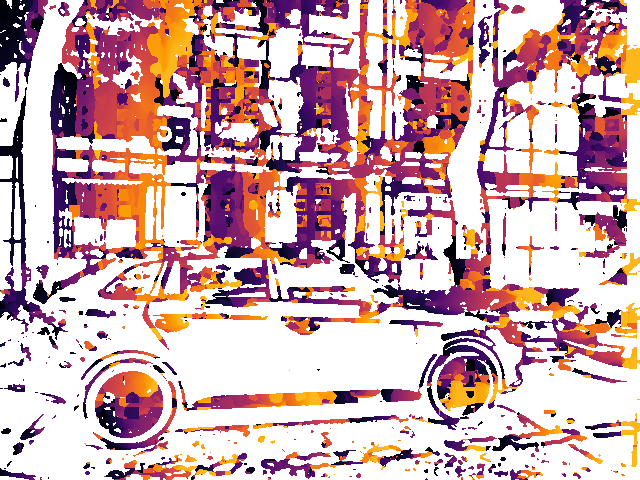}};
        \node [anchor=west] at \sccone {\textcolor{lime}{\bf HSM}};
	\end{tikzpicture} \hspace{-2.2mm}
	\begin{tikzpicture}[spy using outlines={green,magnification=\ssmag,size=\ssizz},inner sep=0]
		\node {\includegraphics[width=\sswidth]{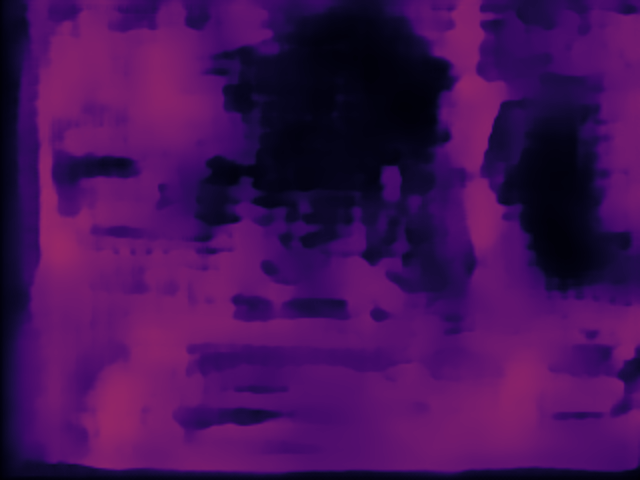}};
        \node [anchor=west] at \sccone {\textcolor{lime}{\bf SIES}};
	\end{tikzpicture} \hspace{-2.2mm}
		\begin{tikzpicture}[spy using outlines={green,magnification=\ssmag,size=\ssizz},inner sep=0]
		\node {\includegraphics[width=\sswidth]{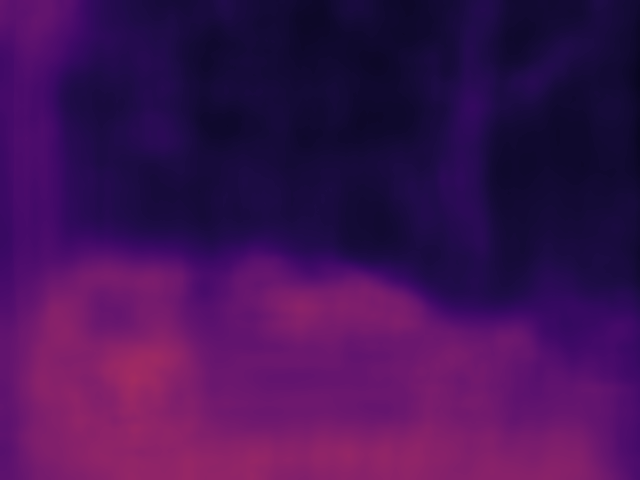}};
        \node [anchor=west] at \sccone {\textcolor{lime}{\bf Our $\mathbf{\hat{D}}_T$}};
	\end{tikzpicture} \hspace{-2.2mm}
		\begin{tikzpicture}[spy using outlines={green,magnification=\ssmag,size=\ssizz},inner sep=0]
		\node {\includegraphics[width=\sswidth]{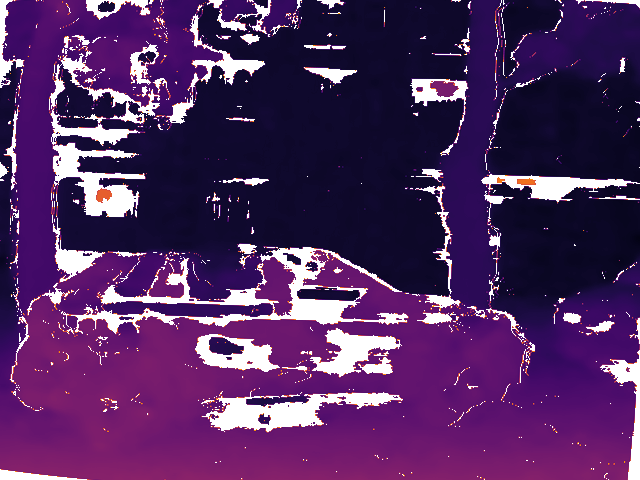}};
        \node [anchor=west] at \sccone {\textcolor{lime}{\bf GT}};
	\end{tikzpicture} \vspace{1mm} \\
	\caption{Qualitative comparison of coarse disparity estimation with the input blurry image and events on the \textit{DSEC-large} (upper two rows) and the \textit{\mydata} (bottom two rows) datasets. To alleviate the modality gap between the intensity image and events for the conventional stereo matching method, \ie, AANet, CFNet, and ACVNet, we introduce E2VID to convert events to the intensity image. HSM is not a CNN-based method and estimates the disparity for the edge image.}
	\label{fig:coarsedisp}
    \vspace{-3mm}
\end{figure*}
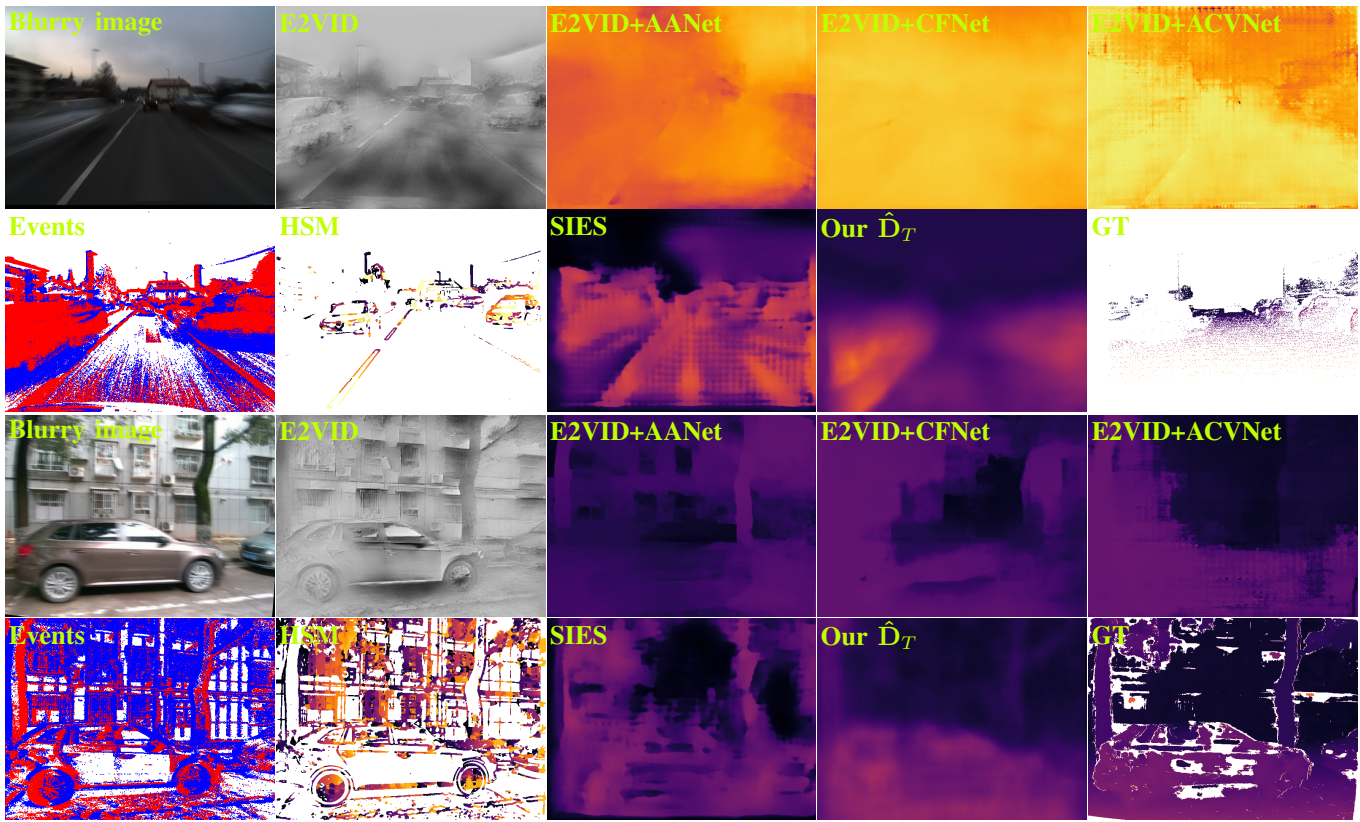

\begin{table}[t]
    \centering
    \caption{Quantitative comparisons on the single disparity map estimation task of the proposed \myname\ to the state-of-the-art methods on the \textit{DSEC-large} and the \textit{StEIC} datasets.}
    \vspace{-2.25mm}
    \begin{tabular}{l c c c c}
        \toprule[1.pt]
        \multirow{2}{*}{Method} & EPE$\downarrow$ & \textgreater1px$\downarrow$ & \textgreater3px$\downarrow$ & \textgreater5px$\downarrow$ \\
        & (px) & (\%) & (\%) & (\%) \\
        \hline
        \textbf{\textit{DSEC-large}}\\
        \hline
        E2VID\cite{rebecq2019high}+AANet\cite{xu2020aanet} & 31.1731 & 99.23 & 97.58 & 95.78 \\
        E2VID\cite{rebecq2019high}+CFNet\cite{shen2021cfnet} & 49.9314 & 99.59 & 98.70 & 97.98 \\
        E2VID\cite{rebecq2019high}+ACVNet\cite{xu2022attention} & 37.5815 & 97.89 & 93.90 & 90.29 \\
        HSM\cite{kim2022real} & 22.8623 & 94.30 & 82.15 & 71.56 \\
        SIES\cite{gu2022self} & 22.1798 & 98.73 & 96.17 & 93.52 \\
        Motion-ETR\cite{zhang2021exposure}+HSM\cite{kim2022real} & 25.4449 & 94.03 & 83.20 & 74.50 \\
        Motion-ETR\cite{zhang2021exposure}+SIES\cite{gu2022self} & 19.3395 & 98.09 & 94.29 & 90.26 \\
        \myname\ (Ours) & \textbf{4.5909} & \textbf{83.70} & \textbf{51.33} & \textbf{27.46} \\
        \hline
        \textbf{\textit{\mydata}}\\
        \hline
        E2VID\cite{rebecq2019high}+AANet\cite{xu2020aanet} & 20.0473 & 97.55 & 92.55 & 87.35 \\
        E2VID\cite{rebecq2019high}+CFNet\cite{shen2021cfnet} & 28.3269 & 97.53 & 92.65 & 87.93 \\
        E2VID\cite{rebecq2019high}+ACVNet\cite{xu2022attention} & 19.8449 & 97.12 & 91.30 & 85.17 \\
        HSM\cite{kim2022real} & 21.0064 & 96.71 & 89.89 & 83.26 \\
        SIES\cite{gu2022self} & 18.8159 & 96.81 & 90.59 & 84.79 \\
        Motion-ETR\cite{zhang2021exposure}+HSM\cite{kim2022real} & 18.7898 & 96.22 & 88.29 & 80.15 \\
        Motion-ETR\cite{zhang2021exposure}+SIES\cite{gu2022self} & 18.8724 & 96.82 & 90.87 & 85.34 \\
        \myname\ (Ours) & \textbf{4.2386} & \textbf{80.07} & \textbf{47.15} & \textbf{26.14} \\
        \toprule[1.pt]
    \end{tabular}
    \label{tab:disp_single}
    \vspace{-3mm}
\end{table}

\begin{table}[t]
    \centering
    \caption{Ablation study for the proposed \myname\ in the sequence prediction task on the \textit{DSEC-large} dataset.}
    \begin{tabular}{p{0.28cm}<{\centering}|p{2.9cm}|p{0.9cm}<{\centering} p{0.9cm}<{\centering} p{0.9cm}<{\centering}}
        \hline
        Ex. & Method & PSNR$\uparrow$ & SSIM$\uparrow$ & LPIPS$\downarrow$\\
        \hline
        1 & w/o DispNet & 28.287 & 0.8622 & 0.1144 \\
        2 & w/ all & \textbf{28.995} & \textbf{0.8759} & \textbf{0.0994} \\
        \hline
    \end{tabular}
    \label{tab:abla0}
    \vspace{-2mm}
\end{table}

\subsection{Results of Disparity Estimation}
\label{sec:result_disp}
From a blurry image and the corresponding events, \myname\ can directly estimate the coarse disparity $\mathbf{\hat{D}}_T$ with DispNet. In this section, we evaluate the performance of \myname\ in the single disparity estimation task. We utilize the widely-used evaluation metrics End-Point Error (EPE) and the percentage of pixels whose disparity errors are greater than 1, 3, and 5 pixels.

We compare the proposed \myname\ with state-of-the-art methods, including three conventional stereo matching methods, \ie, AANet \cite{xu2020aanet}, CFNet \cite{shen2021cfnet} and ACVNet \cite{xu2022attention}, and two disparity estimation methods for stereo event and intensity cameras, \ie, HSM \cite{kim2022real} and SIES \cite{gu2022self}. Existing conventional stereo-matching methods cannot handle the inputs with multiple modalities. Thus, we combine the above three conventional methods with the widely used event-based image reconstruction method E2VID \cite{rebecq2019events}, denoting as E2VID+AANet/CFNet/ACVNet. For two cross-modal stereo matching methods HSM \cite{kim2022real} and SIES \cite{gu2022self}, an individual deblurring module Motion-ETR\cite{zhang2021exposure} is added to alleviate the burden of the blurry images, denoting as Motion-ETR+HSM/SIES.

The quantitative and qualitative results are respectively shown in \cref{tab:disp_single,fig:coarsedisp}. Despite the help of E2VID, conventional stereo matching methods \cite{xu2020aanet, shen2021cfnet, xu2022attention} cannot cover the gap between different modalities, leading to disastrous results. Considering the connection between the intensity and event data via the physical generation model of two cameras, HSM and SIES establish a more satisfactory data association between two views than three conventional stereo-matching methods. However, they still meet the performance drop when facing the intensity image degradation caused by motion blurs. Event with Motion-ETR\cite{zhang2021exposure} as a pre-processing motion deblurring module, HSM and SIES are not competitors of our \myname.

\def\ssxxsone{(0.6,-0.62)} 
\def\ssyysone{(-0.52,0.6)} 
\def\ssxxstwo{(-1.2,.86)} 
\def\ssyystwo{(1.37,-0.58)} 
\def\ssizz{1cm} 
\def\sswidth{0.1645\textwidth} 
\def\ssmag{2.5}
\def\scc{(-1.4,0.92)}
\def\sccone{(-1.4,0.92)}

\begin{figure*}[t]
	\centering
	\begin{tabular}{c c c c c c}
	\hspace{-3mm} \begin{tikzpicture}[spy using outlines={yellow,magnification=\ssmag,size=\ssizz},inner sep=0]
		\node {\includegraphics[width=\sswidth]{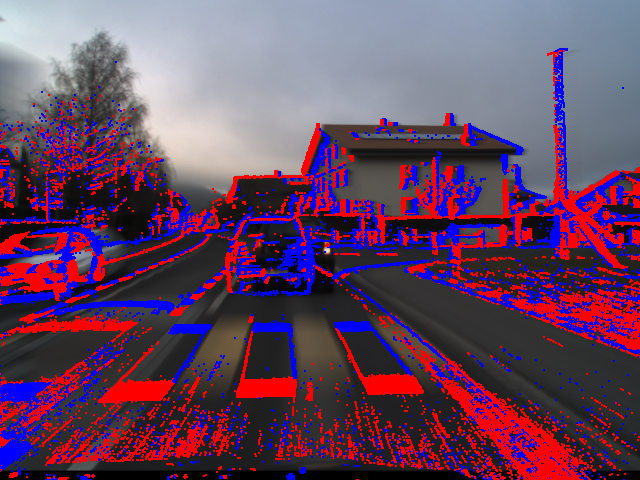}};
		\spy on \ssxxsone in node [left] at \ssyysone;
	\end{tikzpicture}& \hspace{-5.2mm}
	\begin{tikzpicture}[spy using outlines={yellow,magnification=\ssmag,size=\ssizz},inner sep=0]
		\node {\includegraphics[width=\sswidth]{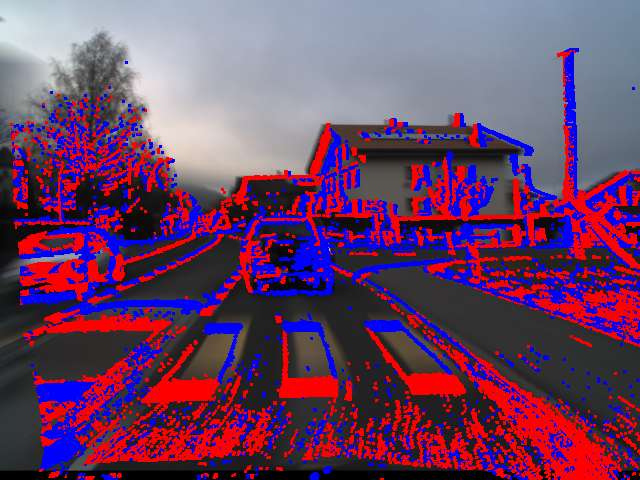}};
		\spy on \ssxxsone in node [left] at \ssyysone;
	\end{tikzpicture}& \hspace{-5.2mm}
	\begin{tikzpicture}[spy using outlines={yellow,magnification=\ssmag,size=\ssizz},inner sep=0]
		\node {\includegraphics[width=\sswidth]{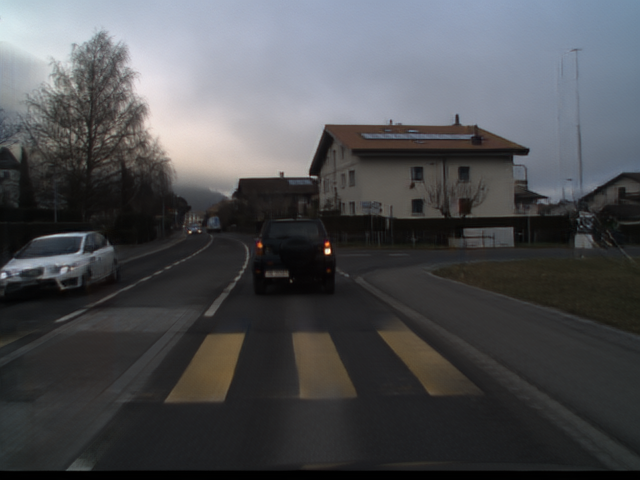}};
		\spy on \ssxxsone in node [left] at \ssyysone;
	\end{tikzpicture}& \hspace{-5.2mm}
		\begin{tikzpicture}[spy using outlines={yellow,magnification=\ssmag,size=\ssizz},inner sep=0]
		\node {\includegraphics[width=\sswidth]{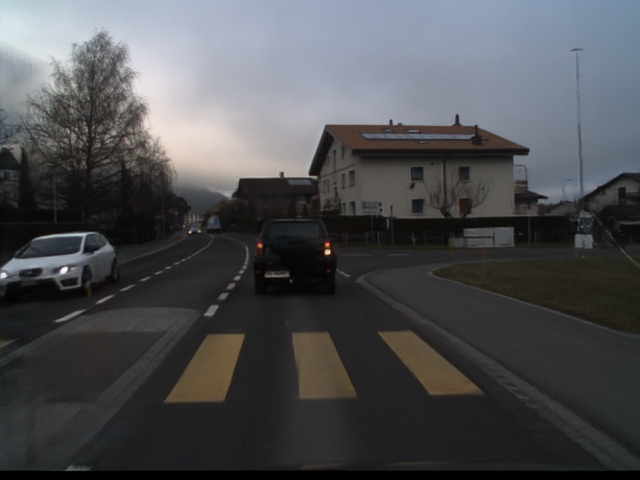}};
		\spy on \ssxxsone in node [left] at \ssyysone;
	\end{tikzpicture}& \hspace{-5.2mm}
		\begin{tikzpicture}[spy using outlines={yellow,magnification=\ssmag,size=\ssizz},inner sep=0]
		\node {\includegraphics[width=\sswidth]{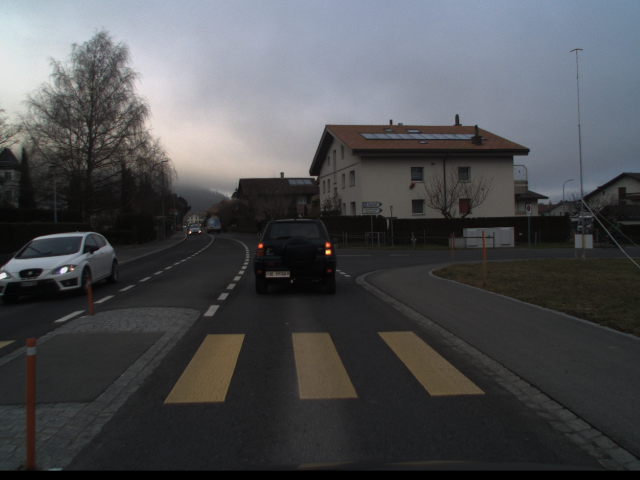}};
		\spy on \ssxxsone in node [left] at \ssyysone;
	\end{tikzpicture}& \hspace{-5.2mm}
		\begin{tikzpicture}[spy using outlines={yellow,magnification=\ssmag,size=\ssizz},inner sep=0]
		\node {\includegraphics[width=\sswidth]{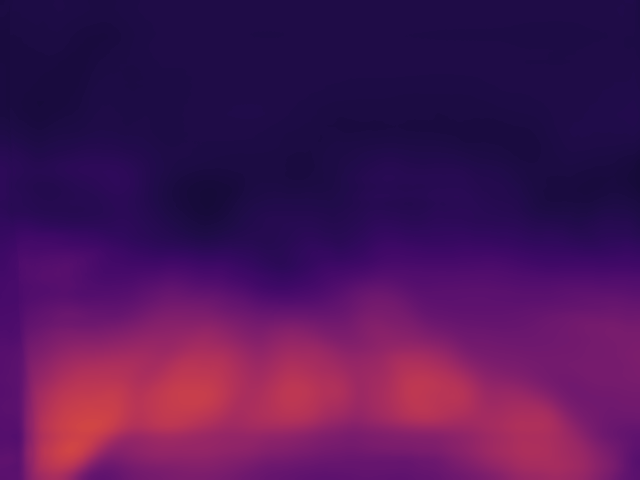}};
	\end{tikzpicture} \\
	\hspace{-3mm} \small{Original data} & \small{Aligned data} & \small{w/o DispNet} & \small{w/ all} & \small{GT} & \small{Predicted disparity}
	\end{tabular}
	\caption{Qualitative ablation study for DispNet on the \textit{DSEC-large} dataset.}
	\label{fig:abladisp}
\end{figure*}

\begin{table}[t]
    \centering
    \caption{Ablation study for DblrNet in the sequence prediction task on the \textit{DSEC-small} dataset.}
    \begin{tabular}{p{0.28cm}<{\centering}|p{0.7cm}<{\centering} p{0.7cm}<{\centering} p{0.7cm}<{\centering}|p{0.9cm}<{\centering} p{0.9cm}<{\centering} p{0.9cm}<{\centering}}
        \hline
        Ex. & DP & BDE & AFF & PSNR$\uparrow$ & SSIM$\uparrow$ & LPIPS$\downarrow$\\
        \hline
        1 & & & & 29.497 & 0.8872 & 0.0882 \\
        2 & \checkmark & & & 30.649 & 0.9078 & 0.0717\\
        3 & \checkmark & & \checkmark & 30.811 & 0.9097 & 0.0709\\
        4 & \checkmark & \checkmark & & 30.941 & 0.9110 & 0.0702\\
        5 & \checkmark & \checkmark & \checkmark & \textbf{31.063} & \textbf{0.9126} & \textbf{0.0692}\\
        \hline
    \end{tabular}
    \label{tab:abla}
    \vspace{-2mm}
\end{table}

\def\ssxxsone{(-0.85,-0.85)} 
\def\ssyysone{(-0.58,0.63)} 
\def\ssxxstwo{(0.3,-0.4)} 
\def\ssyystwo{(-0.58,0.65)} 
\def\ssizz{0.8cm} 
\def\sswidth{0.162\textwidth} 
\def\ssmag{5.5}
\def\scc{(-1.4,0.92)}
\def\sccone{(-1.4,0.92)}

\begin{figure*}[]
	\centering
	\begin{tabular}{c c c c c c}
    \begin{tikzpicture}[spy using outlines={yellow,magnification=\ssmag,size=\ssizz},inner sep=0]
		\node {\includegraphics[width=\sswidth]{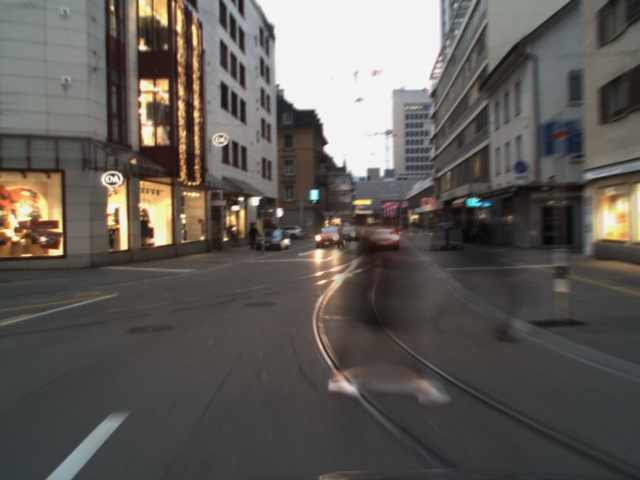}};
		\spy on \ssxxstwo in node [left] at \ssyystwo;
	\end{tikzpicture}& \hspace{-4.8mm}
	\begin{tikzpicture}[spy using outlines={yellow,magnification=\ssmag,size=\ssizz},inner sep=0]
		\node {\includegraphics[width=\sswidth]{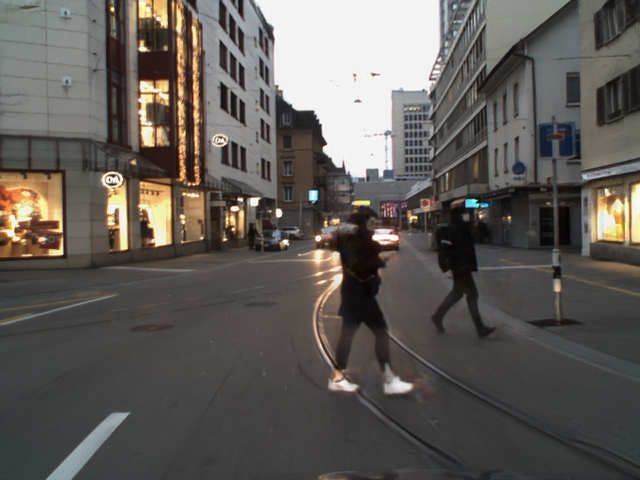}};
		\spy on \ssxxstwo in node [left] at \ssyystwo;
	\end{tikzpicture}& \hspace{-4.8mm}
	\begin{tikzpicture}[spy using outlines={yellow,magnification=\ssmag,size=\ssizz},inner sep=0]
		\node {\includegraphics[width=\sswidth]{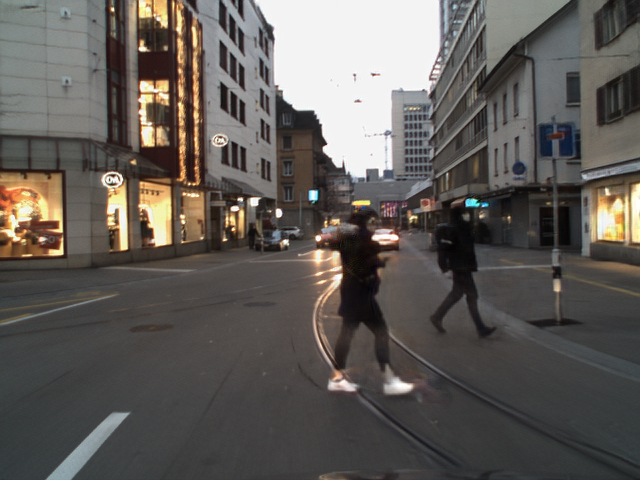}};
		\spy on \ssxxstwo in node [left] at \ssyystwo;
	\end{tikzpicture}& \hspace{-4.8mm}
		\begin{tikzpicture}[spy using outlines={yellow,magnification=\ssmag,size=\ssizz},inner sep=0]
		\node {\includegraphics[width=\sswidth]{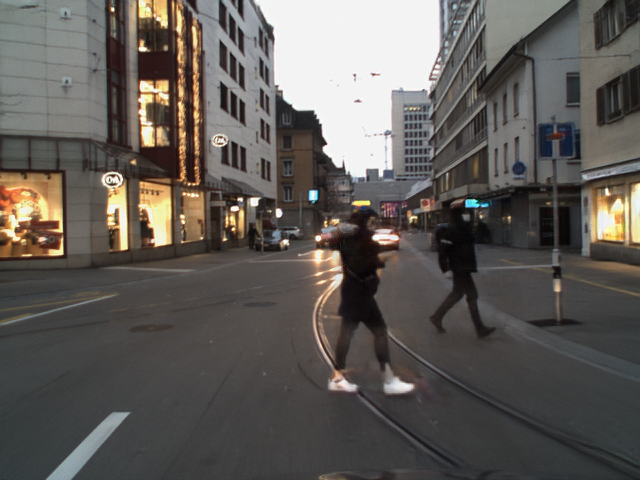}};
		\spy on \ssxxstwo in node [left] at \ssyystwo;
	\end{tikzpicture}& \hspace{-4.8mm}
		\begin{tikzpicture}[spy using outlines={yellow,magnification=\ssmag,size=\ssizz},inner sep=0]
		\node {\includegraphics[width=\sswidth]{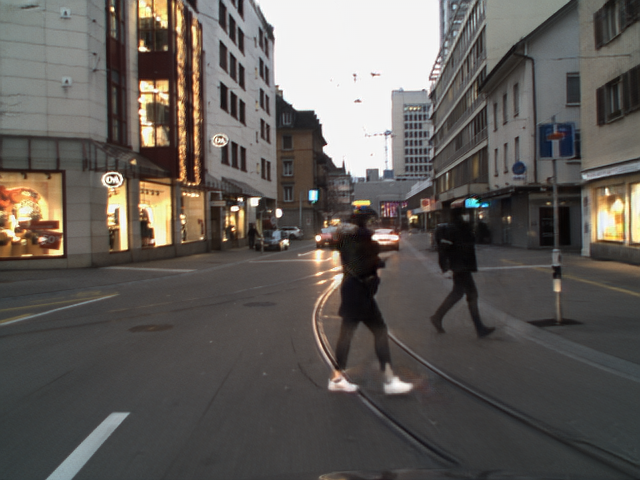}};
		\spy on \ssxxstwo in node [left] at \ssyystwo;
	\end{tikzpicture}& \hspace{-4.8mm}
	\begin{tikzpicture}[spy using outlines={yellow,magnification=\ssmag,size=\ssizz},inner sep=0]
		\node {\includegraphics[width=\sswidth]{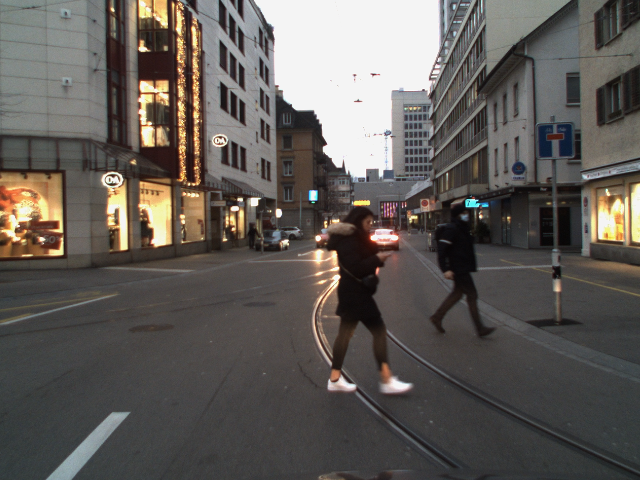}};
		\spy on \ssxxstwo in node [left] at \ssyystwo;
	\end{tikzpicture}\\
	\small{Blurry image} & \small{w/o BDE, AFF} & \small{w/o BDE} & \small{w/o AFF} & \small{w/ all} & \small{GT}
	\end{tabular}
	\caption{Qualitative ablation study for DblrNet on the \textit{DSEC-small} dataset.}
	\label{fig:abla}
\end{figure*}

\subsection{Ablation Study}
In this section, we study the contribution of the several network modules, including DispNet, and Dual Path (DP), BDE, and AFF in DblrNet for the St-ED task. We can draw the following conclusions:

\subsubsection{Necessity of DispNet} \cref{sec:result_disp} has shown that the proposed \myname\, with DispNet, can produce more reliable disparity results than other stereo-matching algorithms without the need for ground-truth depths. We further prove the contribution of DispNet in avoiding the interference of spatially dislocated data and producing clear and sharp deblurring results. 
As shown in Fig. \ref{fig:abladisp} and Tab. \ref{tab:abla0}, DispNet contributes to receding the parallax between the blurry image and the corresponding event streams, which, although not very accurate, can also guide to the pre-alignment between the input events and blurry image from two views.


\subsubsection{Importance of DP, BDE, and AFF} To verify the contribution of each module in DblrNet to the fine matching and fusion of the coarsely aligned data, we construct the \textit{DSEC-small} dataset, where we perform the ablation study for DblrNet alone to avoid interference from DispNet. Specifically, we pair the events from the left event camera “cam0” and the images from the left intensity camera “cam1” with the small baseline of 0.05 meters to form the \textit{DSEC-small} dataset, which can be regarded as the pre-aligned version of the \textit{DSEC-large} dataset.

As demonstrated in Tab. \ref{tab:abla}, the DP approaches are considered as the multi-dimension fusion for misaligned data, outperforming the single-path network (denoted by Ex.1) that simply concatenates the blurry image and events as inputs. As validated by comparing Ex. 4 and 2 (or Ex. 5 and 3), the BDE block can boost the performance for simple or complex feature embedding approaches via aligning cross-modal signals at the pixel level while suppressing artifacts. The AFF block plays an important role in dealing with motion occlusion. In Fig. \ref{fig:abla}, the network without AFF blocks suffers from the overlapping shadows caused by the occlusion regions between the event and the intensity view.

\section{Conclusion}
A novel coarse-to-fine framework named \myname\ for event-based motion deblurring with misaligned event-intensity data has been proposed where the parallax between the event and the intensity view is exploited to alleviate artifact generation caused by the data misalignment at the pixel level. We achieve this by implementing coarse alignment of the multi-view inputs and then finely fusing and enhancing the event and intensity information, where multi-view data with modality discrepancy are effectively leveraged for motion deblurring in real-world scenarios. To this end, the cascaded DDFE modules are proposed for the cross-modal integration and enhancement of the intensity image and events in the feature domain. Furthermore, we build a stereo hybrid camera system to capture a new dataset containing intensity images, dense depths, and real-world event streams. Extensive experiments demonstrate that the proposed \myname\ establishes state-of-the-art performance under the real-world stereo event and intensity camera setup.

\bibliographystyle{IEEEtran}
\bibliography{ref}

\newpage

 
\vspace{11pt}




\vfill

\end{document}